\documentclass{article} 
\usepackage{iclr2027_conference}
\usepackage{newtxtext}

\usepackage{amsmath,amsfonts,bm}

\def\eqref#1{equation~\ref{#1}}

\def\1{\bm{1}}

\DeclareMathAlphabet{\mathsfit}{\encodingdefault}{\sfdefault}{m}{sl}
\SetMathAlphabet{\mathsfit}{bold}{\encodingdefault}{\sfdefault}{bx}{n}

\usepackage{amsmath}
\usepackage{amssymb}
\usepackage{booktabs}
\usepackage{graphicx}
\usepackage{float}
\floatstyle{ruled}
\newfloat{algorithm}{tbp}{loa}
\floatname{algorithm}{Algorithm}
\graphicspath{{figures/}{figures/result_tables/}}
\usepackage{hyperref}
\hypersetup{hidelinks}
\usepackage{microtype}
\usepackage{url}
\usepackage{xcolor}

\newcommand{\method}{\textsc{CrypticFlow}}
\newcommand{\rev}[1]{#1}

\title{\rev{Hierarchical Hybrid-Coordinate Flow Matching for Accurate and Physically Plausible Apo-to-Holo Protein Generation}}

\author{
\small
Minjong Kim$^{1}$, Jaemin Sim$^{1}$, Dongwoo Kim$^{1}$, Myunghyun Jang$^{1}$, Eun-Sol Kim$^{2,\ast}$, Juyong Lee$^{1,\ast}$ \\[4pt]
\normalfont\scriptsize
$^{1}$College of Pharmacy, Seoul National University \qquad $^{2}$Department of Computer Science, Hanyang University \\[2pt]
\normalfont\scriptsize
First author: \texttt{rlaalswhd199@snu.ac.kr} \qquad
$^{\ast}$Correspondence: \texttt{nicole23@snu.ac.kr}, \texttt{eunsolkim@hanyang.ac.kr}
}

\iclrfinalcopy  

\begin{document}

\maketitle
\lhead{Preprint. Under review.}
\raggedbottom

\begin{abstract}
Predicting holo-like protein conformations from apo structures is essential in structure-based drug design. 
Most existing apo-to-holo models operate in Cartesian space and achieve strong global accuracy. 
However, high global accuracy does not ensure physically valid local backbone geometry. 
Internal-coordinate models offer local control and validity, but regressive generation can amplify angular errors---the lever-arm effect. 
We introduce \method{}, a hierarchical flow-matching model that combines these two complementary representations. 
An ESM-3-conditioned E(3)-equivariant graph network first generates sparse
C$\alpha$ anchors in Cartesian space. A conditional midpoint Transformer then
recursively fills the gaps in local internal coordinates. During recursive
generation, the decoder constrains the distance between each midpoint and any
sequence-adjacent parent anchor to the corresponding apo C$\alpha$--C$\alpha$
distance. On the D3PM test split, \method{}
achieves the lowest median and mean C$\alpha$ RMSD at published baseline
capacities. Across five
independent training models, the median and mean direct RMSD are $1.333\pm0.056$~\AA{}
and $2.766\pm0.041$~\AA, respectively. The quality of local geometry also improves
substantially. Our model achieves $99.58\pm0.02\%$ adjacent-distance fidelity
and $95.87\pm0.25\%$ pseudo-angle fidelity, measured within 0.2~\AA{} and
$20^\circ$ of the holo reference. 
Then, we reconstruct the predicted C$\alpha$-only traces into all-atom structures using the cg2all coarse-to-all-atom backmapping method and evaluate their physical and geometric quality with MolProbity. 
Our model achieves a MolProbity score of $3.034\pm0.054$, outperforming the evaluated baselines.
These results show that a hierarchical flow-matching model with hybrid
coordinates can \rev{recover global accuracy while retaining local fidelity}.
\end{abstract}

\section{Introduction}
\label{sec:introduction}

Structure-based drug design (SBDD) depends on the receptor conformation used
for screening. A ligand-bound, or holo, structure reveals a ligand-compatible
pocket. However, obtaining one requires a suitable ligand and an experiment
that captures the bound state. Screening therefore often treats an apo
structure as rigid, which can hide the pocket that a drug must recognize.

This problem is particularly important for \emph{cryptic pockets}. These sites
are not apparent in a static apo structure but can become accessible through
protein motion or a binding-associated conformational change
\citep{vajda2018cryptic,meller2023pocketminer}. Covalent fragments, for example,
revealed the switch-II pocket of KRAS(G12C) \citep{ostrem2013kras}. This
discovery enabled clinically effective inhibitors such as sotorasib and
adagrasib \citep{skoulidis2021sotorasib,janne2022adagrasib}. Predicting
holo-like conformations from apo structures could similarly expand receptor
ensembles before experimental holo structures are available.

Recent methods learn conditional bridges or flows from paired apo--holo
structures
\citep{somnath2023sbalign,nobis2025fdbm}. They are commonly
evaluated using aligned C$\alpha$ RMSD. RMSD measures global accuracy, but it
does not ensure local-geometry fidelity. A prediction may approach the target
while distorting adjacent distances, pseudo-angles, or nonlocal contacts.
These errors become important when the C$\alpha$ trace is reconstructed into
an all-atom structure.

Cartesian and internal coordinates offer complementary advantages. Cartesian
updates represent global motion and work naturally with equivariant networks
\citep{satorras2021egnn}. However, they do not explicitly enforce local
backbone geometry. Internal coordinates represent distances and angles explicitly
\citep{wu2024foldingdiff,li2023int2cart}, but chain-wide generation can suffer
from the lever-arm effect \citep{li2023int2cart,singh2025torsionspace}.
A small angular error can displace many downstream residues.

\rev{We therefore predict sparse Cartesian anchors, then complete
their intervals in a frame defined by two current parents and an apo reference
direction. The main midpoint model uses local internal coordinates and copies
a radius only when its parent is sequence-adjacent. Ablations identify this
copying as a small refinement and the local internal-coordinate state (ICS)
representation as the stronger bias for adjacent and reconstructed bond/angle
fidelity. The shallow hierarchy limits, but does
not eliminate, lever-arm amplification.} Its multiscale design is inspired by Protein Autoregressive Modeling
(PAR) \citep{qu2026par}.

Our contributions are:

\begin{itemize}
    \item \rev{To combine the complementary strengths of Cartesian and
    internal-coordinate representations, we first generate a coarse holo-like
    scaffold as sparse Cartesian anchors with an EGNN, then
    recursively complete the intervening residues with a Transformer in
    anchor-relative local internal coordinates.}

    \item \rev{By representing each midpoint relative to its generated parent
    anchors and the apo reference, the refinement stage introduces an
    inductive bias toward physically plausible local geometry.}

    \item On the fixed 150-pair D3PM test split, \method{} achieves the lowest
    median and mean C$\alpha$ RMSD among the compared methods\rev{, at
    published capacities,} across five
    independently trained models. It also achieves the strongest
    local-geometry fidelity among the evaluated outputs and the lowest
    MolProbity score among the evaluated baselines after cg2all reconstruction.
\end{itemize}
\section{Related Work}
\label{sec:related_work}

\noindent\textbf{Apo-to-holo transport.}
D3PM catalogs protein structural pairs, including apo--holo pairs, with motions
from local residues to whole domains \citep{peng2022d3pm}.
Aligned Diffusion Schr\"odinger Bridges (SBAlign) learns transport from paired
observations \citep{somnath2023sbalign}.
Augmented Bridge Matching (ABM) provides a paired transport formulation used as
a baseline \citep{debortoli2023abm}. Fractional Diffusion Bridge
Models (FDBM) extend this framework from Brownian to approximate fractional
Brownian noise, introducing temporal
correlations \citep{nobis2025fdbm}.
SBAlign, ABM, and FDBM model Cartesian C$\alpha$ coordinates.
Sesame instead applies flow matching to backbone
SE(3) frames, combining rotations and Cartesian translations
\citep{minan2025sesame}.
SBAlign, FDBM, and Sesame evaluate apo-to-holo
prediction on D3PM. 
DynamicFlow generates
ligand-conditioned, full-atom pocket ensembles \citep{zhou2025dynamicflow}.

\noindent\textbf{Equivariant Cartesian models.}
E(n)-equivariant graph neural networks update coordinates with invariant
scalars and relative vectors \citep{satorras2021egnn}. This construction
preserves the transformation law of a 3D structure while learning geometric
messages. ESM-3 is a pretrained
multimodal protein language model (PLM) that learns contextual representations
over protein sequence, structure, and function tokens
\citep{hayes2025esm3}. We use ESM-3 representations precomputed only from
the apo full sequence: residue-level embeddings provide sequence context to
both the anchor and midpoint models, while the beginning-of-sequence (BOS)
embedding provides a global condition to the anchor model.

\noindent\textbf{Transformers for conditional generation.}
Transformers use multi-head self-attention to model interactions among tokens
\citep{vaswani2017attention}. Diffusion Transformers (DiT) adapt Transformer
blocks to generative denoising, with adaptive layer normalization providing
a mechanism for timestep and class conditioning
\citep{peebles2023dit}.

\noindent\textbf{Internal coordinates and multiscale generation.}
FoldingDiff generates protein backbones through torsional variables
\citep{wu2024foldingdiff}. Int2Cart shows that correlations among internal
coordinates affect Cartesian reconstruction quality \citep{li2023int2cart}.
PAR replaces residue-wise autoregression with a sequence of successively finer
protein scales \citep{qu2026par}. We combine these ideas differently. The
top-level anchor state is represented in Cartesian space. At each finer level,
every midpoint is defined relative to two already generated parent anchors in a
local internal-coordinate system. We evaluate the completed C$\alpha$
trace directly. After cg2all backmapping \citep{heo2024cg2all}, we assess the
reconstructed all-atom structure with MolProbity \citep{chen2010molprobity}.

\section{Method}
\label{sec:method}

\subsection{Problem Formulation and Hierarchical Architecture Overview}

\begin{figure}[H]
\centering
\includegraphics[
  page=1,
  trim=10 26 6 84,
  clip,
  width=\linewidth
]{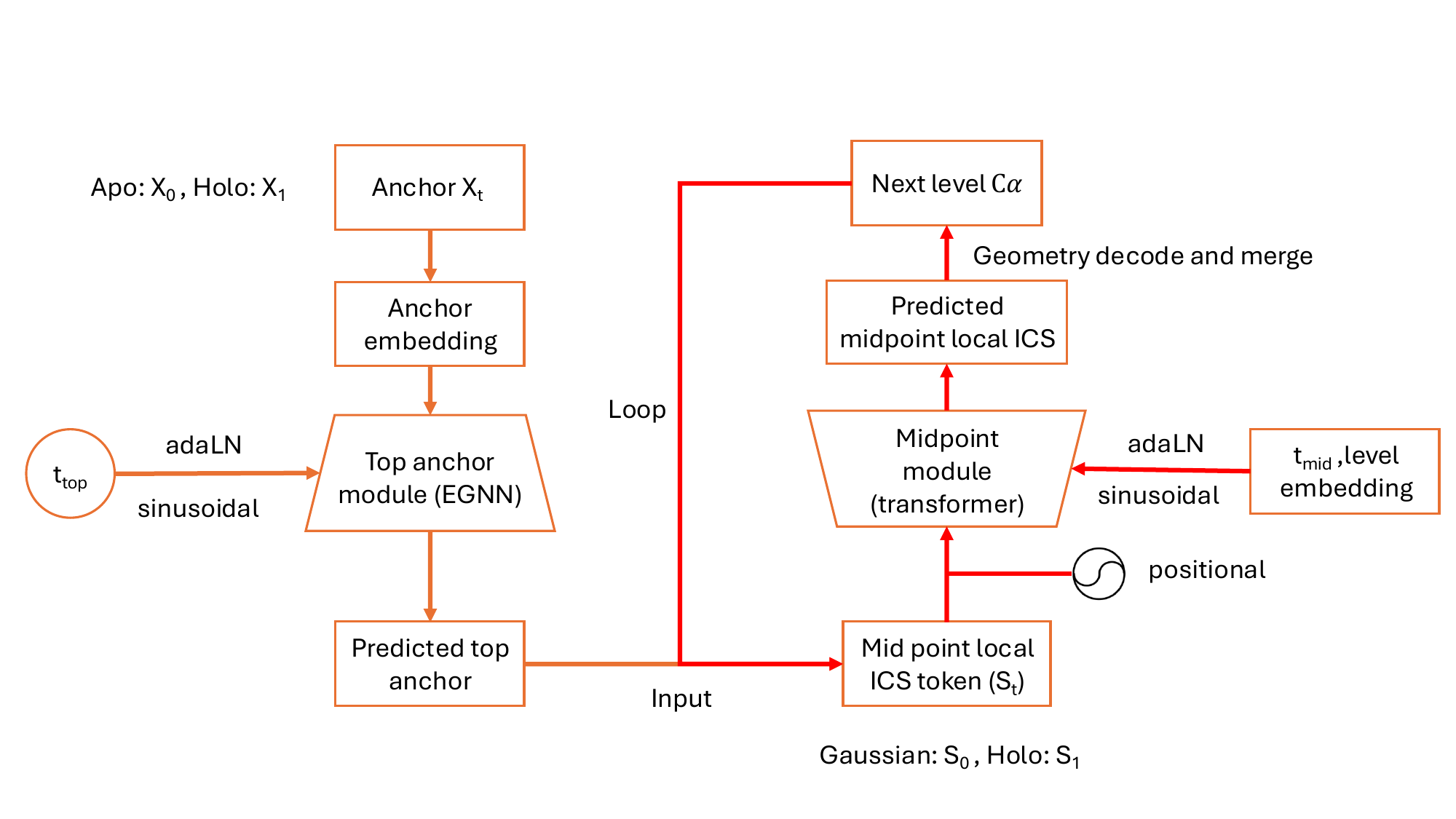}
\caption{Overview of \method{} architecture. The top-anchor module (EGNN)
transports apo anchors to holo-like anchor positions in Cartesian coordinates.
The midpoint module (Transformer) generates midpoints from Gaussian noise in
local internal coordinates defined by their parent anchors and an apo reference
direction. The geometry decoder converts these local internal-coordinate states
into Cartesian midpoint coordinates. Red arrows show the recursive loop: each
decoded midpoint joins the partial trace and serves as an anchor at the next
level until the trace is complete. Training uses paired clean targets and
teacher-forced parent-anchor contexts.}
\label{fig:overall_architecture}
\end{figure}

Let the residue-matched apo and holo C$\alpha$ traces be
$\mathbf{X}_{\mathrm{apo}},\mathbf{X}_{\mathrm{holo}}\in\mathbb{R}^{L\times3}$,
and let the residue features formed by concatenating D3PM descriptors with
full-sequence ESM-3 residue embeddings be $\mathbf{F}$.
The top-anchor EGNN additionally uses the ESM-3 BOS embedding as a global
condition. We learn a conditional model that predicts $\mathbf{X}_{\mathrm{holo}}$ from
$\mathbf{X}_{\mathrm{apo}}$ and $\mathbf{F}$, without ligand input.
\method{} separates global motion from local geometry: a Cartesian flow first
moves sparse anchors, and a local-coordinate flow then fills the gaps between
them. Figure~\ref{fig:overall_architecture} shows the two stages.
The complete training and recursive inference procedure is given in
Appendix~\ref{app:generation_algorithm}.

D3PM DB contains only residues shared by the apo and holo structures.
Removing unmatched residues can introduce gaps in the retained trace, so a
single protein chain may be represented as multiple contiguous segments.
We treat these gaps as chain breaks and do not allow refinement intervals to
cross them. Within each segment, we select the first and last residues,
together with every $s$-th interior residue, as the top anchor set. We use
$s=4$. After generating these anchors,
we split every unresolved interval $[i,j]$ at
$m=\lfloor(i+j)/2\rfloor$, where $\lfloor\cdot\rfloor$ rounds down to
the nearest integer. All midpoints at the same recursion level are generated
together. They are then merged into the partial trace and become
parents at the next level. A full stride-4 interval contains three missing
residues and is completed in two midpoint levels.
Let the top anchor set be $\mathcal A$, the midpoints inserted at level $\ell$
be $\mathcal M_\ell$, and the anchor residues already placed before level
$\ell$ be
$\mathcal P_\ell=\mathcal A\cup\bigcup_{h<\ell}\mathcal M_h$. Thus,
$\mathcal P_\ell$ contains the original top anchors and
all midpoints from earlier levels. Writing $\mathbf{X}$ for a candidate holo trace,
the hierarchy gives
\begin{equation}
p_\theta(\mathbf{X}\mid \mathbf{X}_{\mathrm{apo}},\mathbf{F})
=p_{\mathrm{top}}(\mathbf{X}_{\mathcal A}\mid \mathbf{X}_{\mathrm{apo}},\mathbf{F})
\prod_{\ell=1}^{D}
p_{\mathrm{mid}}(\mathbf{X}_{\mathcal M_\ell}\mid
\mathbf{X}_{\mathcal P_\ell},\mathbf{X}_{\mathrm{apo}},\mathbf{F},\ell),
\label{eq:hierarchical_factorization}
\end{equation}
where the number of refinement levels is $D$. Each midpoint factor is a joint
conditional distribution over a level, not a product of independent residue
predictions. Generation is sequential across levels and parallel within each
level.

\subsection{Clean-Endpoint Conditional Flow}
\label{sec:flow}

We use the linear conditional path of flow matching
\citep{lipman2023flowmatching} and parameterize both networks to predict the
clean endpoint. For a
source state $\mathbf{z}_0$, target state $\mathbf{z}_1$, and conditioning information $\mathbf{c}$,
\begin{equation}
    \mathbf{z}_t=(1-t)\mathbf{z}_0+t \mathbf{z}_1,\qquad t\in[0,1).
    \label{eq:path}
\end{equation}
Instead of predicting velocity directly, each network uses a clean-endpoint
predictor, $f_\theta$, to estimate the target state:
\begin{equation}
    \mathcal{L}_{\mathrm{clean}}
    =\mathbb{E}_{\mathbf{z}_0,\mathbf{z}_1,\mathbf{c},t}\left[\left\|f_\theta(\mathbf{z}_t,t,\mathbf{c})-\mathbf{z}_1\right\|_2^2\right].
    \label{eq:clean_loss}
\end{equation}
For the path in Equation~\ref{eq:path}, this prediction defines a velocity
field
\begin{equation}
    \widehat{\mathbf v}_\theta(\mathbf{z}_t,t,\mathbf{c})
    =\frac{f_\theta(\mathbf{z}_t,t,\mathbf{c})-\mathbf{z}_t}{\max(1-t,\epsilon)}.
    \label{eq:clean_velocity}
\end{equation}
Here the positive numerical stabilizer is $\epsilon>0$. The anchor flow starts from apo
anchor coordinates and targets their holo coordinates. The midpoint flow
starts from Gaussian noise and targets a normalized holo local-coordinate state
defined in Section~\ref{sec:midpoint}. The networks thus share a training
principle but use different representations and source distributions.

\subsection{Coarse Cartesian Anchor Generation}
\label{sec:top_egnn}

\begin{figure}[H]
\centering
\includegraphics[
  page=3,
  trim=13 247 13 364,
  clip,
  width=\linewidth
]{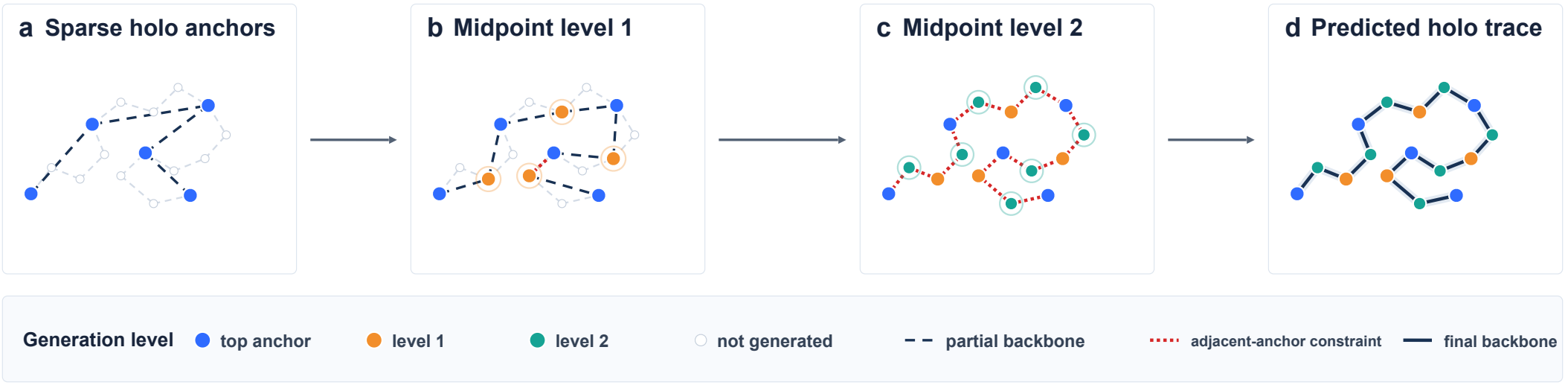}
\caption{Hierarchical coarse-to-fine C$\alpha$ generation. A Cartesian flow
first predicts sparse stride-4 holo anchors. The midpoint Transformer inserts one
midpoint in each active parent interval at each recursive level. Two midpoint
levels fill full stride-4 intervals, while adjacent-anchor distance constraints
help preserve local geometry in the completed trace.}
\label{fig:hierarchical_generation}
\end{figure}

The top-anchor EGNN predicts clean holo anchor coordinates from $\mathbf{X}_t$ in eight layers. Initial
node features combine D3PM descriptors and ESM-3 residue
embeddings; a shared condition combines flow time with the ESM-3 BOS
embedding. For each central anchor $i$, the graph takes the union of
neighbors $j$ selected by apo-space proximity, current-state proximity, and sparse-sequence adjacency.
The neighbor list is fixed within each forward pass through the eight-layer
top-anchor EGNN, while current-state distances and geometric edge features
are recomputed at every EGNN layer. During inference, the graph is
rebuilt at each Euler step.

At each layer, the network decides how each neighbor should influence an
anchor's movement. For anchor $i$ and neighbor $j$, it uses the vector from
$i$ to $j$ in the current coordinates,
$\mathbf r_{ij}^{\mathrm{cur}}=\mathbf{X}_j^{(k)}-\mathbf{X}_i^{(k)}$, and the
corresponding vector in the apo structure,
$\mathbf r_{ij}^{\mathrm{apo}}=\mathbf{X}_{\mathrm{apo},j}-\mathbf{X}_{\mathrm{apo},i}$.
At layer $k$, their soft-normalized forms,
$\bar{\mathbf r}_{ij}^{\mathrm{cur}}$ and
$\bar{\mathbf r}_{ij}^{\mathrm{apo}}$, are
\begin{equation}
\bar{\mathbf r}_{ij}^{\mathrm{cur}}=
\frac{\mathbf r_{ij}^{\mathrm{cur}}}{\|\mathbf r_{ij}^{\mathrm{cur}}\|+1},\qquad
\bar{\mathbf r}_{ij}^{\mathrm{apo}}=
\frac{\mathbf r_{ij}^{\mathrm{apo}}}{\|\mathbf r_{ij}^{\mathrm{apo}}\|+1}.
\end{equation}
The corresponding current and apo distances are
$d_{ij}^{\mathrm{cur}}=\|\mathbf r_{ij}^{\mathrm{cur}}\|$ and
$d_{ij}^{\mathrm{apo}}=\|\mathbf r_{ij}^{\mathrm{apo}}\|$. Each raw distance
$d$ becomes a soft-normalized magnitude $d/(d+1)<1$; this transformation smoothly
suppresses near-coincident pairs and bounds long-range contributions.
At each layer, an edge MLP forms an invariant message $\mathbf{m}_{ij}^{(k)}$ from the conditioned
node features of nodes $i$ and $j$ and their edge geometry. This message determines a sigmoid
gate $q_{ij}^{(k)}$, which controls the edge contribution, and signed weights
$a_{ij}^{(k)},b_{ij}^{(k)}$, which scale the two soft-normalized vectors.
Averaging these contributions over the neighbor set
$\mathcal N(i)$ gives the coordinate update:
\begin{equation}
\mathbf{X}_i^{(k+1)}=\mathbf{X}_i^{(k)}+
\frac{1}{|\mathcal N(i)|}
\sum_{j\in\mathcal N(i)}q_{ij}^{(k)}
\left(a_{ij}^{(k)}\bar{\mathbf r}_{ij}^{\mathrm{cur}}+
b_{ij}^{(k)}\bar{\mathbf r}_{ij}^{\mathrm{apo}}\right).
\label{eq:coord_update}
\end{equation}
The weights are invariant to a joint rigid transformation of the current and
apo structures, while the relative vectors transform with them. The update
therefore follows the same transformation: it does not depend on the chosen
coordinate frame. Therefore, every coordinate-update layer is
$\mathrm{E}(3)$-equivariant. If the input structures are jointly translated, rotated, or reflected, the output transforms
in the same way.
Appendix~\ref{app:equivariance} evaluates the composed generator under
random proper rotations and translations and separately tests reflection.

After eight layers, let the anchor-validity mask be $m_i\in\{0,1\}$ and the
final displacement of anchor $i$ be
$\mathbf{D}_i=\mathbf{X}_i^{(8)}-(\mathbf{X}_t)_i$. The network removes the
masked mean displacement,
$\bar{\mathbf D}=\sum_i m_i\mathbf{D}_i/\sum_i m_i$, and returns the clean
anchor estimate,
$\widehat{\mathbf X}_{1,i}=(\mathbf{X}_t)_i+\mathbf{D}_i-\bar{\mathbf D}$,
for each valid anchor. This suppresses
uniform translation while preserving relative conformation and the current
anchor centroid. Appendix~\ref{app:top_details} details the top-anchor EGNN architecture,
graph construction, and node and coordinate updates.

\subsection{\texorpdfstring{\rev{Local Internal Coordinates and Geometry Decoding}}{Local Internal Coordinates and Geometry Decoding}}
\label{sec:local_ics}

\begin{figure}[H]
\centering
\includegraphics[width=\linewidth]{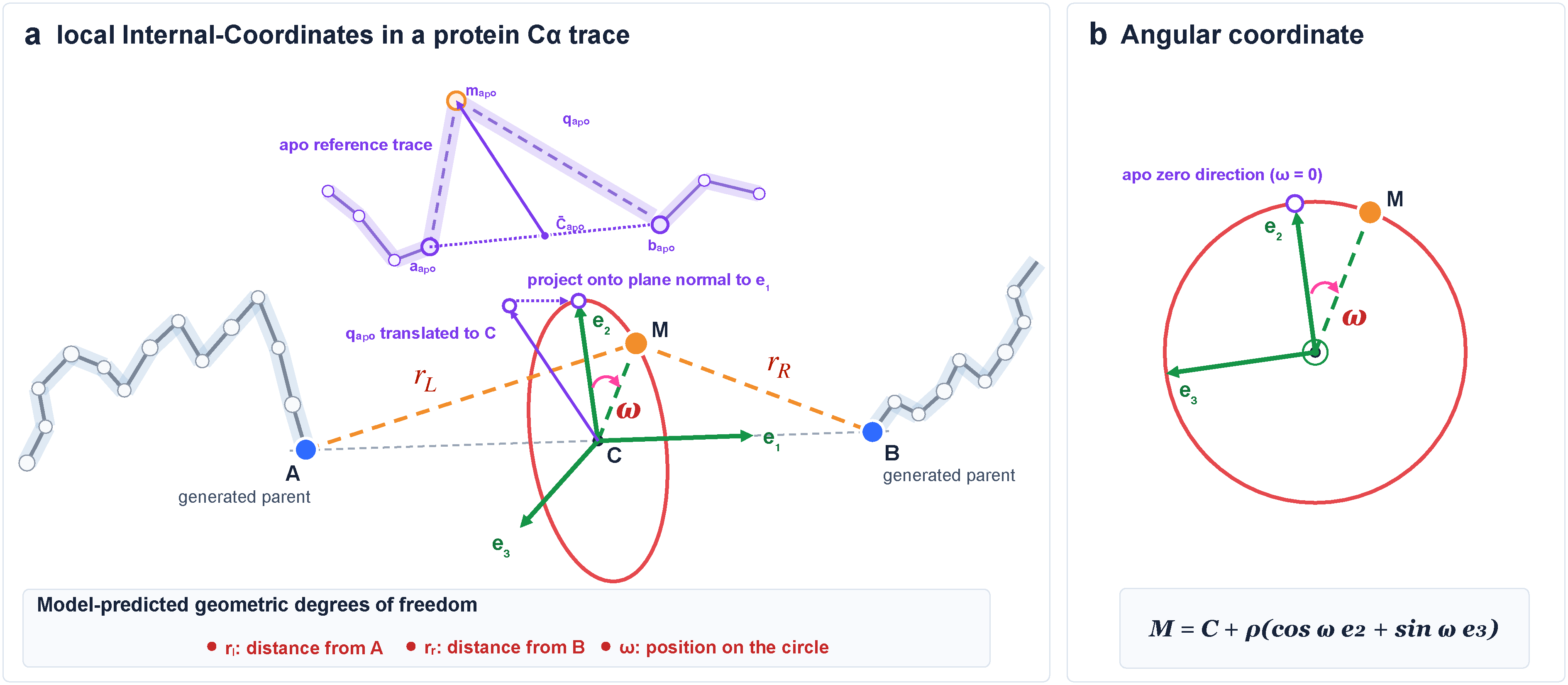}
\caption{Local internal-coordinate geometry. (a) Parent anchors $\mathbf{A}$ and
$\mathbf{B}$ define a sphere-intersection circle. The apo midpoint direction $\mathbf{q}_{\mathrm{apo}}$ is translated to $\mathbf{C}$ and projected onto the plane normal to $\mathbf{e}_1$ to define
$\mathbf{e}_2$; $\mathbf{e}_3$ completes the local frame. Predicted $r_L$, $r_R$, and $\omega$
place $\mathbf{M}$ on the circle. (b) Viewed along $+\mathbf{e}_1$, $\omega$ is measured from
the apo-defined $+\mathbf{e}_2$ direction. Sequence-adjacent radii are fixed to the
corresponding apo C$\alpha$--C$\alpha$ distances, reducing the predicted
degrees of freedom from three to two or one.}
\label{fig:local_ics_geometry}
\end{figure}

Let the left and right parent anchors be $\mathbf{A}$ and $\mathbf{B}$, and let
their generated sequence midpoint be $\mathbf{M}$. Rather than predict
an unconstrained Cartesian displacement, we represent $\mathbf{M}$ by its parent
distances $r_L=\|\mathbf{M}-\mathbf{A}\|$, $r_R=\|\mathbf{M}-\mathbf{B}\|$, and an angle $\omega$ around the
parent axis. For distinct parents, define
\begin{equation}
d=\|\mathbf{B}-\mathbf{A}\|,\quad
\mathbf{e}_1=\frac{\mathbf{B}-\mathbf{A}}{d},\quad
\eta=\frac{d^2+r_L^2-r_R^2}{2d},\quad
\mathbf{C}=\mathbf{A}+\eta \mathbf{e}_1.
\label{eq:local_circle}
\end{equation}
Here the signed axial distance from $\mathbf{A}$ is $\eta$, the
sphere-intersection circle center is $\mathbf{C}$, and the parent-axis direction
is $\mathbf{e}_1$. The circle radius is
$\rho=\sqrt{\max(r_L^2-\eta^2,0)}$. When the two distance constraints are
feasible, their intersection is a circle centered at $\mathbf{C}$ with radius $\rho$.

The apo trace supplies the reference direction for $\omega$. The apo
coordinates of the residues represented by $\mathbf{A},\mathbf{B},\mathbf{M}$
are $\mathbf{A}_{\mathrm{apo}},\mathbf{B}_{\mathrm{apo}},\mathbf{M}_{\mathrm{apo}}$,
respectively. In particular,
$\mathbf{M}_{\mathrm{apo}}$ is the observed apo position of the sequence-midpoint
residue, whereas $\mathbf{M}$ is its generated position. We denote the
apo parents' Cartesian midpoint by
$\bar{\mathbf C}_{\mathrm{apo}}=(\mathbf{A}_{\mathrm{apo}}+\mathbf{B}_{\mathrm{apo}})/2$ and define
\begin{equation}
\mathbf{q}_{\mathrm{apo}}=\mathbf{M}_{\mathrm{apo}}-\bar{\mathbf C}_{\mathrm{apo}},\qquad
\mathbf{e}_2=\frac{(\mathbf{I}-\mathbf{e}_1\mathbf{e}_1^\top)\mathbf{q}_{\mathrm{apo}}}
{\|(\mathbf{I}-\mathbf{e}_1\mathbf{e}_1^\top)\mathbf{q}_{\mathrm{apo}}\|},\qquad
\mathbf{e}_3=\mathbf{e}_1\times \mathbf{e}_2.
\label{eq:local_frame}
\end{equation}
Here the $3\times3$ identity matrix is $\mathbf I$, the apo reference vector is
$\mathbf q_{\mathrm{apo}}$, the zero-angle direction is $\mathbf e_2$, and the
third frame direction is $\mathbf e_3$. Geometrically, we move the tail of
$\mathbf{q}_{\mathrm{apo}}$ to $\mathbf{C}$ without changing
its direction or length, then project it onto the plane through $\mathbf{C}$ with
normal $\mathbf{e}_1$. Normalizing the projected vector gives $\mathbf{e}_2$, the zero-angle
direction on the circle.
If the projected apo vector has zero or near-zero norm, it cannot provide a
stable direction for $\mathbf{e}_2$. We instead project a deterministically selected
least-aligned fallback axis onto the same plane and normalize the result.
Appendix~\ref{app:frame_fallback} explains this construction and the limitation
of using fixed Cartesian axes. The Cartesian midpoint is
\begin{equation}
\mathbf{M}=\mathbf{C}+\rho(\cos\omega\,\mathbf{e}_2+\sin\omega\,\mathbf{e}_3).
\label{eq:local_decode}
\end{equation}
For fixed parents, an angular change moves only this midpoint along its local
circle.
Figure~\ref{fig:hierarchical_generation} visualizes the two-level stride-4
refinement, and Figure~\ref{fig:local_ics_geometry} illustrates the frame
used for local internal coordinates.
Appendix~\ref{app:local_geometry} derives the circle and local frame step by
step, including why translating the apo reference vector to $\mathbf{C}$ and
projecting it onto the circle plane is equivalent to removing its axial
component.

\paragraph{Adjacent-distance constraints.}
Adjacent C$\alpha$--C$\alpha$ distances are tightly constrained by backbone
connectivity and change little between matched apo and holo conformations, as
quantified in Appendix~\ref{app:adjacent_distance_stats}.
\rev{The main configuration replaces a learned radius spanning residues $i$
and $i+1$ with the corresponding apo distance as a fixed local constraint.}
Let the sequence gaps from midpoint $m$ to its left and right parents $i,j$ be
$n_L=m-i$ and $n_R=j-m$, respectively. If either gap is one, the corresponding
radius is fixed to the sanitized apo C$\alpha$--C$\alpha$ distance:
\begin{equation}
n_L=1\Rightarrow r_L=d_{\mathbf{A},\mathbf{M}}^{\mathrm{apo}},\qquad
n_R=1\Rightarrow r_R=d_{\mathbf{M},\mathbf{B}}^{\mathrm{apo}}.
\label{eq:adjacent_constraint}
\end{equation}
Here the sanitized apo distances for the left and right adjacent pairs are
$d_{\mathbf A,\mathbf M}^{\mathrm{apo}}$ and
$d_{\mathbf M,\mathbf B}^{\mathrm{apo}}$, respectively.
Sanitation retains finite apo distances in $[2.5,5.0)$~\AA{} and
replaces other values with 3.80~\AA.
A midpoint therefore follows a \emph{3/2/1-DOF policy}, with three, two, or one
learned geometric variables when zero, one, or both parent radii are fixed.
Across all 1,591 D3PM pairs finds that 99.8202\% of 577,291
within-segment consecutive C$\alpha$ pairs change by less than 0.2~\AA{}
between apo and holo structures
(Appendix~\ref{app:adjacent_distance_stats}).

\paragraph{Sphere feasibility.}
The radii and parent separation must satisfy
\begin{equation}
|r_L-r_R|\leq d\leq r_L+r_R.
\label{eq:sphere_feasibility}
\end{equation}
For an infeasible non-final midpoint, the predicted spheres do not intersect;
we keep the parent anchors fixed and minimally adjust only the learnable radii
to form a valid intersection circle. If Equation~\ref{eq:adjacent_constraint}
fixes one radius to an apo adjacent distance, only the other is adjusted. At a
final leaf, both radii are fixed to apo adjacent distances and remain
unchanged. If the two fixed radii are incompatible with the generated parents,
their spheres do not intersect. The local-ICS geometry decoder then sets the intersection-circle
radius to zero and places the midpoint at $\mathbf{C}=\mathbf{A}+\eta \mathbf{e}_1$ on the parent axis.
This finite numerical fallback is not
a feasibility repair, and the query remains marked infeasible.
Appendix~\ref{app:sphere_repair} gives the exact case-wise corrections. 
In five-seed free test rollouts, $1.882\pm0.486\%$ of midpoint queries are
raw-infeasible before non-final repair; Appendix~\ref{app:feasibility_rate}
reports the seed-wise counts and final-leaf events.

\subsection{Recursive Midpoint Generation}
\label{sec:midpoint}

The midpoint Transformer generates the local states for all active intervals at a level.
Each midpoint local state, $\mathbf{z}^{\mathrm{mid}}\in\mathbb{R}^4$, contains
the left and right radial channels, $u_L,u_R$, and the angular channels,
$\sin\omega,\cos\omega$, so
$\mathbf{z}^{\mathrm{mid}}=(u_L,u_R,\sin\omega,\cos\omega)$.
The midpoint Transformer outputs local-ICS channels, whose radial components
are mapped to bounded radii, while the angular pair avoids the discontinuity at
$\pm\pi$. To balance the scales of the four local-ICS channels, flow matching
is performed on the channel-wise standardized state,
$\widetilde{\mathbf z}=(\mathbf{z}-\boldsymbol{\mu}_{\mathrm{train}})/\boldsymbol{\sigma}_{\mathrm{train}}$,
where the training-target channel mean and standard deviation are
$\boldsymbol{\mu}_{\mathrm{train}}$ and $\boldsymbol{\sigma}_{\mathrm{train}}$.
Before geometry decoding, predicted states are transformed
back to raw local coordinates.

Each query concatenates its noisy local state, parent-distance features, apo
local state, D3PM residue features, and ESM-3 embeddings. Each query token
receives a sinusoidal residue-position embedding. Flow-time and recursion-level
embeddings condition each Transformer block through adaptive layer
normalization (AdaLN). A four-channel head predicts the clean local state;
architectural details appear in Appendix~\ref{app:midpoint_details}.

At each level, the local-ICS geometry decoder maps the generated local internal-coordinate states
to Cartesian midpoint coordinates and merges them with the existing parent anchors
to update the
partial trace. The updated trace supplies the parent anchors for the next
level, and this generate--decode--merge cycle repeats recursively until the
C$\alpha$ trace is complete.

\subsection{Training and Inference}

\paragraph{Training.}
Paired apo--holo structures provide clean targets for both flows. Midpoint
training uses teacher-forced parent contexts; inference instead uses the
parents generated at earlier levels.
The anchor clean-endpoint loss, $\mathcal L_{\mathrm{top}}$, and midpoint
clean-endpoint loss, $\mathcal L_{\mathrm{mid}}$, apply the objective to anchor
coordinates and local midpoint states, respectively. Radius channels fixed by
Equation~\ref{eq:adjacent_constraint} are excluded from the midpoint state
loss. The full objective is
\begin{equation}
\mathcal{L}=\mathcal{L}_{\mathrm{top}}+\mathcal{L}_{\mathrm{mid}}
+\alpha_s\!\left(0.2\mathcal{L}_{\mathrm{xyz}}+0.1\mathcal{L}_{\mathrm{adj}}
+0.1\mathcal{L}_{\mathrm{sparse}}+0.05\mathcal{L}_{\mathrm{feas}}\right).
\label{eq:total_loss}
\end{equation}
The auxiliary losses penalize errors in decoded midpoint coordinates,
$\mathcal L_{\mathrm{xyz}}$; adjacent distances, $\mathcal L_{\mathrm{adj}}$;
sparse pair distances, $\mathcal L_{\mathrm{sparse}}$; and sphere
infeasibility, $\mathcal L_{\mathrm{feas}}$. The auxiliary-loss scale,
$\alpha_s$, increases
linearly from zero to one over the first 500 optimization steps. Flow times
are drawn in equal proportions from $t=0$,
the five-point Euler grid, and a continuous uniform distribution on $[0,1)$.
Appendix~\ref{app:loss_details} gives the exact masks, reductions, and
Smooth-L1 definitions used by every objective term.

\paragraph{Inference.}
The anchor flow first transforms apo anchor coordinates into holo-like anchors.
These anchors initialize a partial trace that is recursively completed by the
midpoint flow. Unlike teacher-forced training, inference builds the hierarchy
from its own predictions: each decoded midpoint is added to the partial trace
and becomes a fixed parent anchor for the next level. Each flow uses five
Euler steps, and evaluation draws one prediction per test protein.

\section{Experimental Setup}
\label{sec:experiments}

\paragraph{Data.}
We use the fixed D3PM-Large split introduced by SBAlign: 1,291 apo--holo pairs
for training, 150 for validation, and 150 for testing
\citep{somnath2023sbalign,peng2022d3pm}. We center and Kabsch-align each pair
during preprocessing \citep{kabsch1976rotation}; segment labels prevent
midpoint generation across mask-defined gaps.

\paragraph{Full-sequence ESM-3 features.}
We run ESM-3 on the full apo SEQRES sequence and gather embeddings for observed
D3PM residues, without holo input. Appendix~\ref{app:esm_mapping} gives the
windowing, residue-mapping, and validation details.

\paragraph{Model and training.}
Complete architecture settings are reported in
Appendix~\ref{app:implementation}. We train for 1,000 epochs with AdamW
\citep{loshchilov2019adamw}, an
effective batch size of eight, and top and midpoint learning rates of
$10^{-4}$ and $5\times10^{-4}$. For each seed in $\{1,3,42,88,99\}$, we
evaluate the checkpoint with the lowest validation mean Kabsch RMSD and report
mean $\pm$ sample standard deviation. Appendices~\ref{app:ablation}
and~\ref{app:compute} report model selection and computational cost.

\paragraph{Baselines and protocol.}
For global RMSD, we use published D3PM results for SBAlign and Sesame and
FDBM-reported results for EGNN, ABM, and FDBM
\citep{somnath2023sbalign,minan2025sesame,nobis2025fdbm}. Local and all-atom
analyses use stored predictions for direct-regression EGNN, SBAlign, ABM, and
FDBM on the same 150 test identifiers; Sesame is omitted because reproducible
per-target outputs were unavailable. We report one prediction from each of five
training seeds.

\paragraph{Metrics.}
Following the D3PM protocol \citep{somnath2023sbalign,peng2022d3pm}, we measure
RMSD directly in the pre-aligned frame without a prediction-to-target fit and
report success below 2, 5, and 10~\AA. Local fidelity uses 0.2~\AA{} distance
and $20^\circ$ pseudo-angle thresholds; C$\alpha$ clashes are non-adjacent
pairs below 3~\AA{} per million eligible pairs.

\paragraph{Relaxation-free all-atom validation.}
Fixed-C$\alpha$ cg2all reconstructs all predictions, and MolProbity scores the
unrelaxed outputs \citep{heo2024cg2all,chen2010molprobity}. All reconstructions
succeed; GT-holo applies the same pipeline to reference C$\alpha$ traces.

\section{Results}
\label{sec:results}

\begin{figure}[H]
\centering
\includegraphics[width=0.76\linewidth]{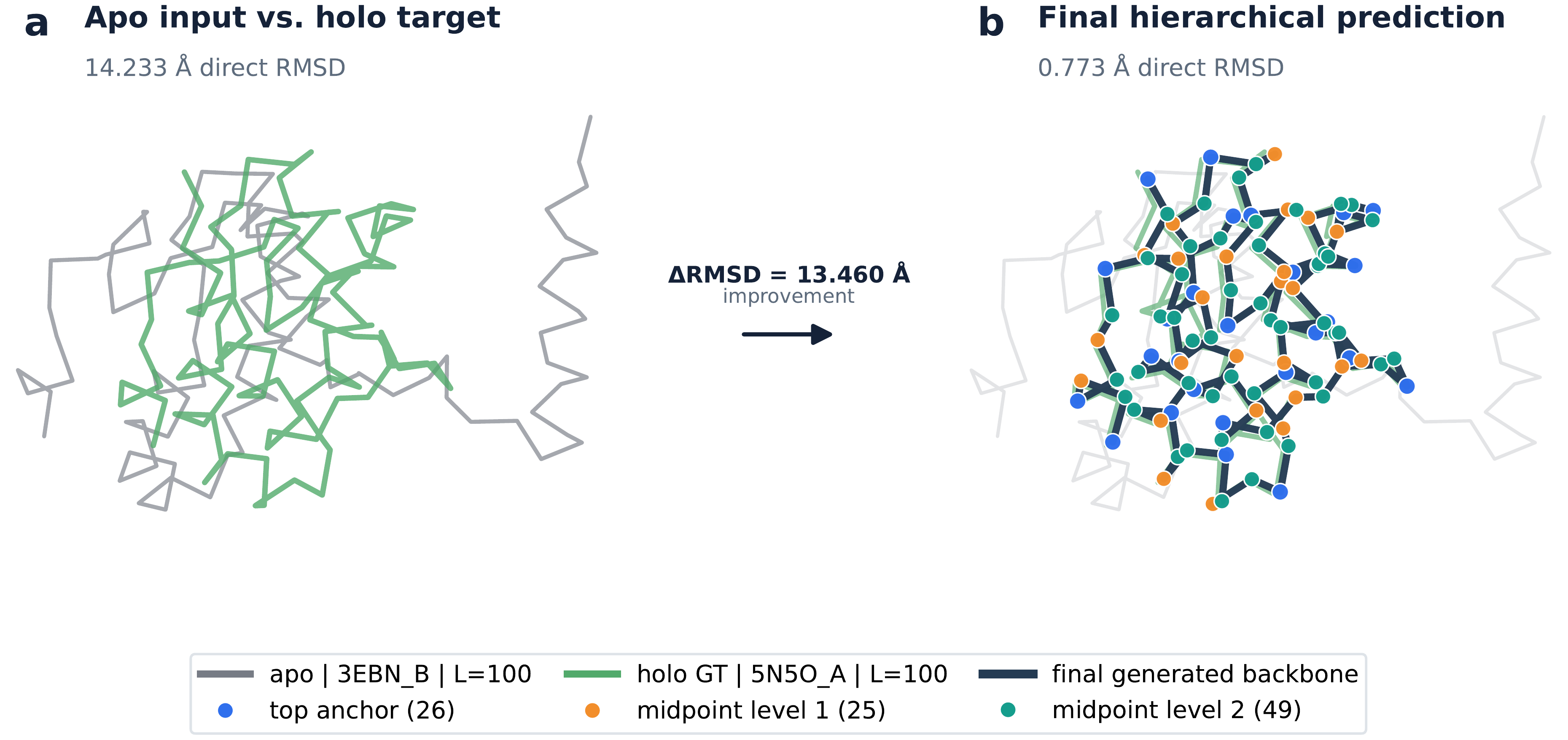}
\caption{Example for the 100-residue pair 3EBN\_B--5N5O\_A. (a) Apo input
(gray) and holo target (green), with 14.233~\AA{} direct C$\alpha$ RMSD.
(b) The prediction (dark blue) reaches 0.773~\AA{} RMSD; markers show 26 top
anchors (blue), 25 level-1 midpoints (orange), and 49 level-2 midpoints (teal).}
\label{fig:structure_visualization}
\end{figure}

Figure~\ref{fig:structure_visualization} visualizes how sparse anchors and two
midpoint levels resolve a large apo--holo displacement.
A corresponding hierarchical inference animation is provided in the accompanying code
archive under \texttt{videos/crypticflow\_hierarchical\_inference}.

\subsection{Apo-to-Holo Accuracy}
\begin{table}[H]
\caption{C$\alpha$ accuracy on the fixed D3PM test split. Ours is mean $\pm$
sample standard deviation over five models; baseline values are taken from
the cited reports \citep{somnath2023sbalign,minan2025sesame,nobis2025fdbm}.
Test std. is the dispersion across test proteins; our row
summarizes this statistic across five models.}
\label{tab:accuracy}
\centering
\begin{tabular*}{\linewidth}{@{\extracolsep{\fill}}lrrrrrr@{}}
\toprule
& \multicolumn{3}{c}{RMSD (\AA) $\downarrow$}
& \multicolumn{3}{c}{Success rate (\%) $\uparrow$}\\
\cmidrule(lr){2-4}\cmidrule(lr){5-7}
Method & Median & Mean & Test std. & $<2$~\AA{} & $<5$~\AA{} & $<10$~\AA{}\\
\midrule
EGNN (reported) & 19.99 & 21.37 & 8.21 & 1.0 & 1.0 & 3.0\\
SBAlign (10, 10) & 3.80 & 4.98 & 3.95 & 0.0 & 69.0 & 93.0\\
SBAlign (100, 100) & 3.81 & 5.02 & 3.96 & 0.0 & 70.0 & 93.0\\
Sesame & 2.87 & 3.65 & 2.95 & 38.0 & 82.0 & 96.0\\
ABM & 2.40 & 3.49 & 3.54 & 43.0 & 84.0 & 96.0\\
FDBM ($H=0.3$) & 2.33 & 3.42 & 3.42 & 43.0 & 85.0 & 97.0\\
FDBM ($H=0.2$) & 2.12 & 3.34 & 3.59 & 48.0 & \textbf{86.0} & 96.0\\
FDBM ($H=0.1$) & 2.20 & 3.44 & 3.57 & 46.0 & 83.0 & 97.0\\
\midrule
\textbf{\method{} (ours)}
& \shortstack{$\mathbf{1.333}$\\[-1pt]$\mathbf{\pm 0.056}$}
& \shortstack{$\mathbf{2.766}$\\[-1pt]$\mathbf{\pm 0.041}$}
& \shortstack{$3.646$\\[-1pt]$\pm 0.026$}
& \shortstack{$\mathbf{58.53}$\\[-1pt]$\mathbf{\pm 0.73}$}
& \shortstack{$84.40$\\[-1pt]$\pm 1.12$}
& \shortstack{$96.53$\\[-1pt]$\pm 0.30$}\\
\bottomrule
\end{tabular*}
\end{table}

\subsection{\texorpdfstring{C$\alpha$}{C-alpha} and All-Atom Structural Validity}
\begin{table}[H]
\caption{Core validity metrics. \method{} reports mean \ensuremath{\pm} sample standard deviation over five models; baselines are single stored predictions.}
\label{tab:physical}
\centering
\begin{tabular*}{\linewidth}{@{\extracolsep{\fill}}lrrr@{}}
\toprule
Method & Dist. $\leq0.2$~\AA{} (\%) $\uparrow$
& Angle $\leq20^\circ$ (\%) $\uparrow$
& MP score $\downarrow$\\
\midrule
GT-holo C$\alpha$ control & 100.0000 & 100.0000 & 2.291\\
\midrule
Direct-regression EGNN & 8.7252 & 37.8666 & 4.861\\
SBAlign & 11.5265 & 46.1836 & 4.624\\
FDBM & 32.0681 & 82.1678 & 4.085\\
ABM & 30.9050 & 82.4563 & 4.089\\
\textbf{\method{} (ours)}
& $\mathbf{99.5825\!\pm\!0.0200}$
& $\mathbf{95.8652\!\pm\!0.2543}$
& $\mathbf{3.034\!\pm\!0.054}$
\\
\bottomrule
\end{tabular*}
\end{table}

Tables~\ref{tab:accuracy} and~\ref{tab:physical} show gains in both global and
local quality. Relative to FDBM ($H=0.2$), \method{} lowers median/mean direct
RMSD from 2.12/3.34~\AA{} to $1.333\pm0.056$/$2.766\pm0.041$~\AA{} and raises
sub-2~\AA{} success from 48.0\% to $58.53\pm0.73\%$. It also achieves
$99.58\pm0.02\%$ distance fidelity, $95.87\pm0.25\%$ angle fidelity, and the
lowest learned-model MolProbity score ($3.034\pm0.054$). Detailed metrics appear in
Appendix~\ref{app:physical_details}. \rev{Controls that isolate the source of
this fidelity, the same comparison at matched baseline capacity, and an
external benchmark are reported in Appendices~\ref{app:test_controls},
\ref{app:capacity} and~\ref{app:cryptobench}.}

\section{Discussion and Limitations}
\label{sec:discussion}

\paragraph{\rev{Where each result comes from.}}
\rev{\method{} splits the problem deliberately, and controls let us say which
stage carries which result. Global accuracy is set almost entirely at the
Cartesian anchor stage. Replacing the generated anchors with ground-truth ones
lowers the mean RMSD from 2.766 to 0.775~\AA, and the midpoint stage adds
almost no error of its own, so 72\% of the remaining global error sits in the
anchors (Appendix~\ref{app:exposure}). Local geometry comes from the
internal-coordinate midpoints: a local-Cartesian midpoint is slightly better on
RMSD but loses the adjacent-distance and reconstructed bond/angle fidelity
(Appendix~\ref{app:mechanistic_controls}). One component we expected to matter
does not. The apo radius copied for a sequence-adjacent parent can be removed
altogether and the distance fidelity is unchanged to within $0.02$~pp
(Appendix~\ref{app:test_controls_radii}). That number is produced by the
representation, not by the constraint.}

\paragraph{\rev{The hierarchy is a position on a trade-off.}}
\rev{Neither pure representation is good on both axes, and each fails where the
other succeeds. A hierarchy-free stride-1 Cartesian flow reaches a lower median
RMSD than \method{} in all five seeds, by $0.180\pm0.037$~\AA, but its
adjacent-distance fidelity is 67.3\% against our 99.6\%. Full-chain
internal-coordinate rollouts reach a higher pseudo-angle fidelity,
98.78--98.83\% against our 95.87\%, but their median RMSD is more than five
times ours, and retraining them at our parameter count closes only about a
tenth of that gap (Appendix~\ref{app:hierarchy_fairness}). \method{} sits
between the two: it gives up $0.180$~\AA{} of median RMSD and gains
$+32.2\pm2.3$~pp of adjacent-distance fidelity. The claim we can make is
therefore not that the hierarchy improves global accuracy, nor that it
preserves local geometry better than any alternative. It is that of these three
settings only the hierarchy is good on both axes at once.}

\paragraph{\rev{What capacity matching changes, and what it does not.}}
\rev{At 34.08M parameters \method{} is substantially larger than the published
bridge baselines, so the main comparison is not capacity matched. We retrained
all four at matched capacity with the same pretrained features. ABM then
attains a lower global RMSD than \method{} on this split, 2.547 against
2.766~\AA{} (Appendix~\ref{app:capacity}), so the accuracy claim holds at the
published baseline capacities rather than in general. Matching capacity does
not change the geometry result: the best capacity-matched baseline still
reaches only 38.5\% adjacent-distance fidelity against our 99.6\%. Size is also
not bought with time, since \method{} remains faster per target than all three
bridge baselines under the hardware-controlled comparison in Appendix
Table~\ref{tab:compute}. Our accuracy result therefore depends on model size,
and our geometry result does not.}

\paragraph{\rev{What transfers to an unseen benchmark.}}
\rev{We then asked whether the geometry result survives a change of
distribution, using CryptoBench, an external cryptic-pocket benchmark, with no
retraining. Every capacity-matched baseline there is worse than leaving the
apo structure untouched, and \method{} is the only evaluated generative model
that improves on it, in both global and pocket C$\alpha$ RMSD
(Appendices~\ref{app:cryptobench} and~\ref{app:cryptobench_pocket}). The
margin is small and the docking gains stay well short of the reconstructed
ground-truth ceiling, so this is evidence that the hybrid representation
survives a distribution shift, not a claim that cryptic-pocket prediction is
solved. Two limits bound it further. \method{} returns a single ligand-free
C$\alpha$ trace, and repeated sampling does not change that: 20 samples of one
target differ by a median of $0.020\pm0.003$~\AA, and an oracle over them
performs the same as a single sample (Appendix~\ref{app:diversity}). It is a
deterministic predictor rather than an ensemble generator, and it never sees
the ligand that opens the pocket.}

\section{Conclusion}
\label{sec:conclusion}

\rev{We introduced \method{}, a hierarchical flow-matching model for
apo-to-holo generation. It places sparse anchors with a Cartesian flow and
fills the gaps between them with recursive local internal coordinates, so that
no dependency path runs longer than two refinements. The design answers a
trade-off. Cartesian models are globally accurate but geometrically
implausible, and internal-coordinate models are the reverse. On the fixed
150-pair D3PM split, \method{} is the only one of these three settings that is
good on both axes, and that holds in each of our five independently trained
models.}

\rev{Our accuracy result and our geometry result do not rest on the same
footing. The global-accuracy lead holds at the published baseline capacities,
but not once every baseline is retrained at our parameter count. The local-geometry and all-atom advantage is
untouched by that matching, and it is the one that transfers: on CryptoBench,
\method{} is the only evaluated generative model that improves on the unmoved
apo input. We read the hybrid representation, rather than the accuracy ranking,
as the result worth carrying forward.}

\rev{What the model still cannot do is return more than one structure.
Repeated sampling gives back the same trace, so \method{} does not address the
conformational heterogeneity that makes cryptic pockets hard. Ligand-conditioned,
all-atom, multi-state generation is the natural next step, and the hierarchical
hybrid-coordinate representation is what we would build it on. The same
representation could also carry over to conformational change driven by
protein--protein interactions or post-translational modifications.}

\subsection*{Reproducibility statement}

Sections~\ref{sec:method} and~\ref{sec:experiments} specify the fixed split,
representations, model dimensions, training objective, inference procedure,
and evaluation metrics. Appendix~\ref{app:implementation} records the complete
main configuration and seed protocol; Appendices~\ref{app:ablation},
\ref{app:mechanistic_controls}, \ref{app:lever_arm}, and~\ref{app:esm_mapping}
document model selection, post-selection controls, perturbation diagnostics,
and sequence mapping. The accompanying code archive contains
the implementation, fixed split manifest, five sanitized training/evaluation
configuration pairs, per-target metric records, and a standard-library
script that recomputes the \method{} rows in Tables~\ref{tab:accuracy}
and~\ref{tab:physical}. Its \texttt{README.md} and
\texttt{REPRODUCIBILITY.md} document the environment, data schema, training,
inference, and all-atom evaluation procedures. Upstream D3PM structures,
ESM-3 assets, and large checkpoints are not redistributed; the archive
documents their preparation and the commands needed to retrain or evaluate a
checkpoint. The archive will be released publicly upon publication.

\subsection*{AI use statement}

Generative AI tools, including OpenAI Prism and Codex, assisted with literature
discovery; manuscript drafting, editing, and translation; method
implementation and software review; dataset validation and formatting;
experimental design; figure and artifact development; and interpretation of
results. They were not used to generate synthetic datasets, formulate or prove
mathematical claims, or propose scientific hypotheses; qualitative and
thematic data analysis was not applicable. The authors manually verified the
cited literature, implementation, experimental artifacts, numerical claims,
and final wording, and take responsibility for the complete content of the
submission.

\bibliography{crypticflow_arxiv}
\bibliographystyle{iclr2027_conference}

\appendix

\section*{Supplementary Benchmark Visualizations}

\begin{figure}[H]
\centering
\includegraphics[width=0.92\linewidth]{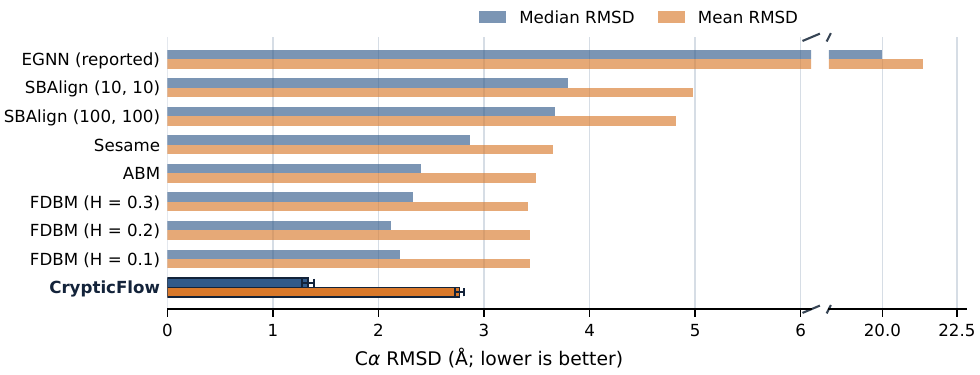}
\caption{Median and mean direct C$\alpha$ RMSD on the fixed D3PM test split
after the dataset's apo-to-holo preprocessing alignment. The broken axis
retains legacy EGNN; \method{} error bars show sample standard deviations over
five models. Baseline values are taken from the cited reports.}
\label{fig:rmsd_benchmark}
\end{figure}

\begin{figure}[H]
\centering
\includegraphics[width=0.70\linewidth]{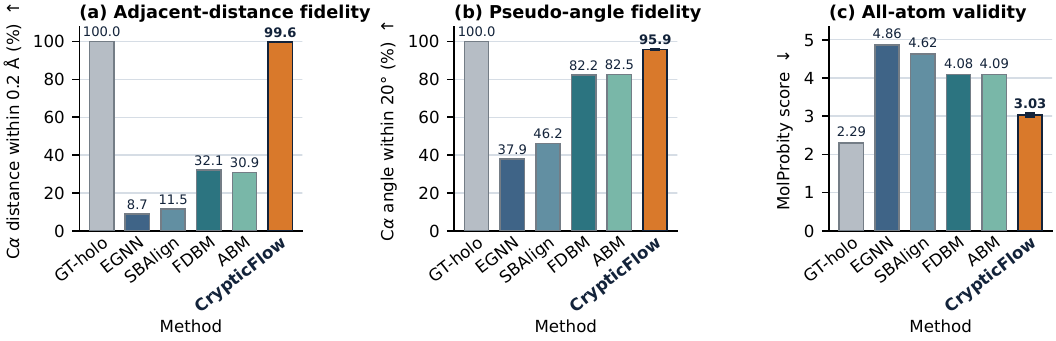}
\caption{Local and all-atom validity. Error bars show sample standard
deviations over five \method{} models. Sesame is omitted because reproducible
per-target predictions were unavailable.}
\label{fig:molprobity_benchmark}
\end{figure}

Figures~\ref{fig:rmsd_benchmark} and~\ref{fig:molprobity_benchmark} visualize
the results reported in main Tables~\ref{tab:accuracy}
and~\ref{tab:physical}, respectively.

\section{Detailed Network Architectures}
\label{app:detailed_architectures}

\subsection{Anchor Network Details}
\label{app:top_details}

Each anchor receives a 145-dimensional D3PM descriptor and a 1,536-dimensional
ESM-3 residue representation. A projection produces the initial invariant
node state $\mathbf{H}^{(0)}\in\mathbb{R}^{|\mathcal A|\times320}$. A sinusoidal time
embedding and the projected ESM-3 BOS representation produce a
320-dimensional condition $\mathbf{c}$, broadcast to all anchors.

The graph unites three directed neighborhoods: 32-nearest neighbors in the
apo structure, 16-nearest neighbors in the flow state $\mathbf{X}_t$ at the start of
the forward pass, and adjacent entries in the sparse anchor sequence whose
original residue index gap is at most 16. Graph indices stay fixed during
that pass. Layer $k$ recomputes the invariant edge features,
$\mathbf{G}_{ij}^{(k)}$,
\begin{equation}
\mathbf{G}_{ij}^{(k)}=\operatorname{RBF}(\|\mathbf{X}_j^{(k)}-\mathbf{X}_i^{(k)}\|)
\,\Vert\,
\operatorname{RBF}(\|\mathbf{X}_{\mathrm{apo},j}-\mathbf{X}_{\mathrm{apo},i}\|)
\,\Vert\,
\operatorname{Fourier}(p_j-p_i),
\end{equation}
with 32, 32, and 17 channels, respectively. Here the original residue index of
anchor $i$ is $p_i$, and the concatenation operator is $\Vert$.

\begin{figure}[t]
\centering
\includegraphics[page=1,width=\linewidth]{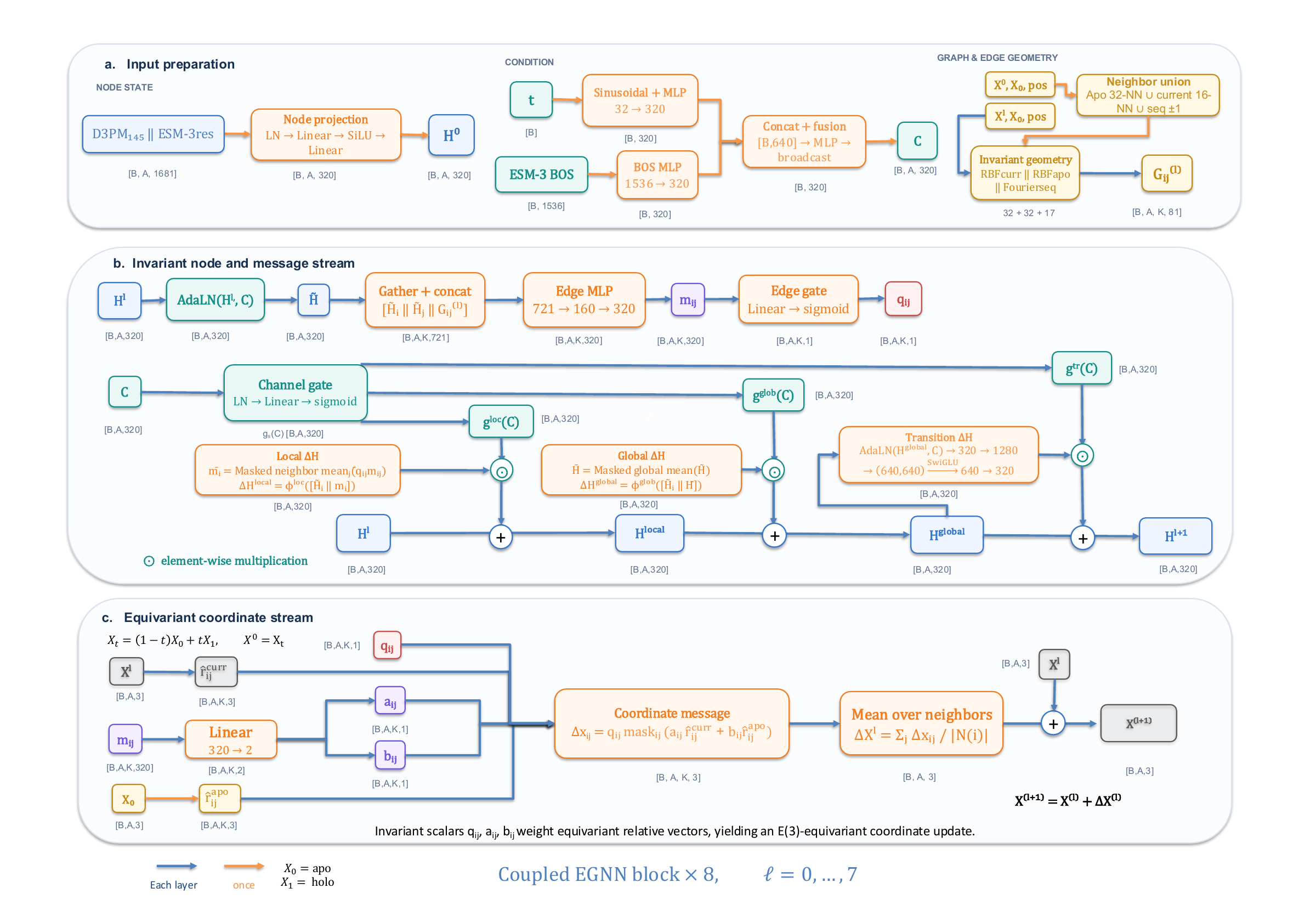}
\caption{Detailed top-anchor Cartesian flow. Precomputed ESM-3 residue and
BOS features condition an eight-layer EGNN. A fixed neighbor union supports
invariant messages and equivariant coordinate updates. Orange paths are
prepared once; blue paths are recomputed in each layer. Tensor dimensions are
shown below the blocks.}
\label{fig:top_architecture_detail}
\end{figure}

Each of eight EGNN layers computes the conditioned node state,
$\widetilde{\mathbf H}_i^{(k)}$; edge message,
$\mathbf m_{ij}^{(k)}$; and message gate, $q_{ij}^{(k)}$:
\begin{align}
\widetilde{\mathbf H}_i^{(k)}
&=\operatorname{AdaLN}(\mathbf{H}_i^{(k)},\mathbf{c}),\\
\mathbf{m}_{ij}^{(k)}
&=\phi_e\!\left(
[\widetilde{\mathbf H}_i^{(k)}\Vert\widetilde{\mathbf H}_j^{(k)}\Vert \mathbf{G}_{ij}^{(k)}]
\right),\\
q_{ij}^{(k)}&=\sigma\!\left(\phi_q(\mathbf{m}_{ij}^{(k)})\right).
\end{align}
The 320-dimensional edge message is multiplied by $q_{ij}^{(k)}$ and averaged
over valid neighbors. Node features receive a local message update, a
masked-global-mean update, and a SwiGLU transition. Each residual is modulated
channel-wise by the condition:
\begin{equation}
\mathbf{H}\leftarrow \mathbf{H}+g_s(\mathbf{c})\odot\Delta \mathbf{H}_s,
\qquad s\in\{\mathrm{local},\mathrm{global},\mathrm{tr}\}.
\label{eq:node_update}
\end{equation}
Here $\Delta\mathbf H_s$ is the candidate update from branch $s$, and
$g_s(\mathbf c)$ is its condition-dependent channel gate.
The edge messages also produce the coordinate weights in
Equation~\ref{eq:coord_update}. Figure~\ref{fig:top_architecture_detail} shows the complete layer-level data flow.

Reading the figure from left to right, D3PM and ESM-3 residue features first
initialize $\mathbf{H}^{(0)}$, while flow time and the ESM-3 BOS embedding produce the
condition $\mathbf{c}$. Current and apo coordinates define the neighbor union and edge
geometry. In every EGNN layer, AdaLN conditions the node features; the center
feature, neighbor feature, and $\mathbf{G}_{ij}^{(k)}$ enter the edge MLP to form
$\mathbf{m}_{ij}^{(k)}$ and the gate $q_{ij}^{(k)}$. The gated messages update node
features through successive local, global, and transition residuals. In
parallel, the same message predicts signed weights for the soft-normalized
current and apo relative vectors, and the gated mixture updates the coordinates.
The updated node states and coordinates enter the next layer. Neighbor indices
stay fixed, whereas current-state geometry is recomputed at each layer. After
eight layers, the final displacement is mean-centered to produce the clean
holo-like anchor estimate.

\subsection{Midpoint Network Details}
\label{app:midpoint_details}

For each active interval, the input token concatenates the noisy local state
(4 channels), current parent-distance features (17), apo local state (4), and
D3PM plus ESM-3 residue features (1,681). Parent-distance features comprise a
normalized distance and 16 Gaussian RBF channels. Layer normalization and a
linear map project the 1,706 input channels to 256, after which a sinusoidal
residue-position embedding is added.

\begin{figure}[t]
\centering
\includegraphics[page=2,width=\linewidth]{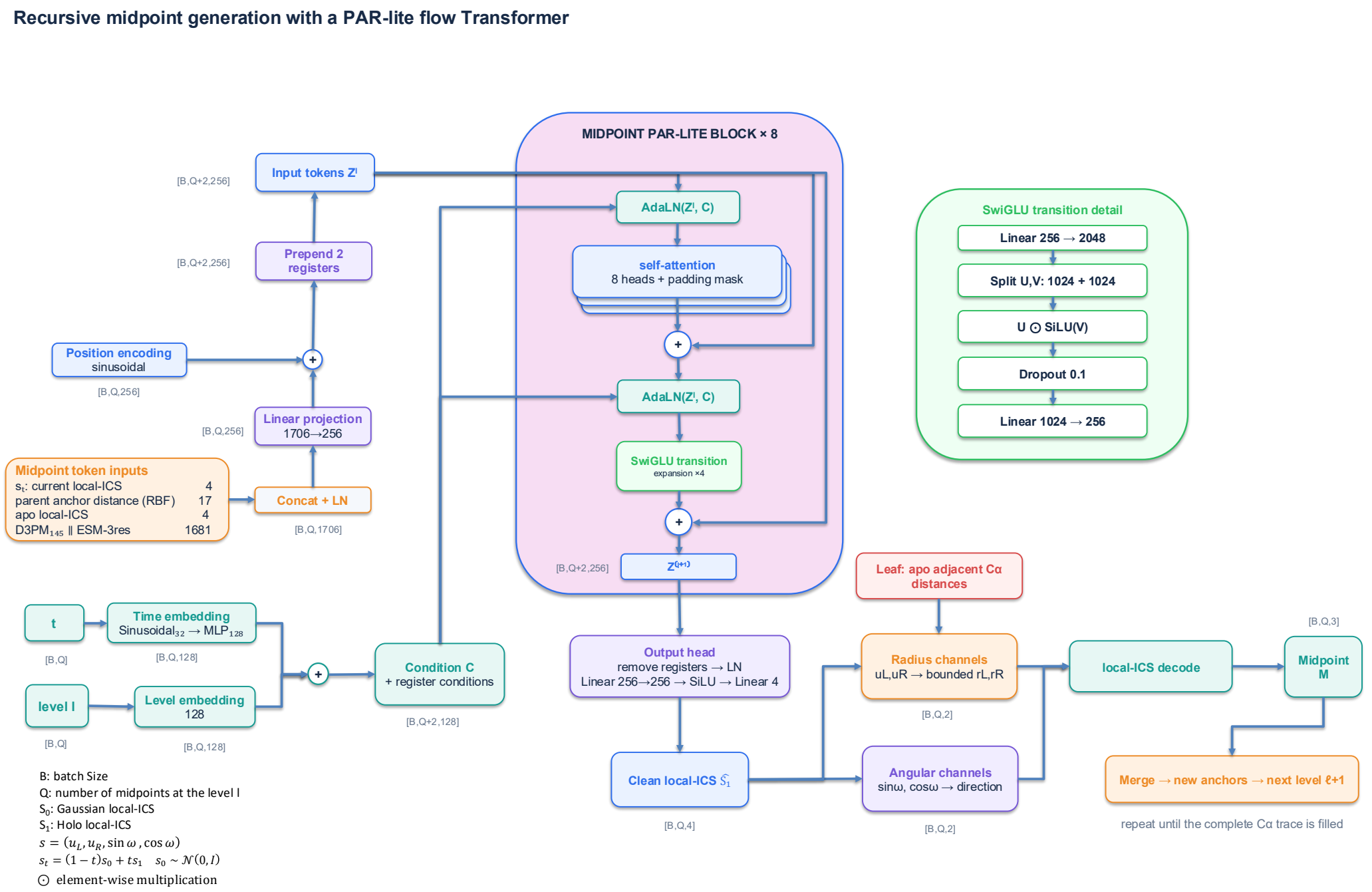}
\caption{Detailed midpoint flow at recursion level $\ell$. The $Q$ active
midpoint queries are processed in parallel with two learned registers.
Self-attention is non-causal. After clean-state integration, bounded radii and
an angular direction are decoded in local internal coordinates and merged into
the trace.}
\label{fig:midpoint_architecture_detail}
\end{figure}

Two learned register tokens are prepended. A sinusoidal flow-time embedding
and a learned recursion-level embedding form the 128-dimensional condition.
Eight non-causal Transformer blocks apply AdaLN-conditioned eight-head
self-attention and an AdaLN-conditioned SwiGLU transition, each with a
residual connection. The registers are removed before a four-channel head
predicts the clean local state. Figure~\ref{fig:midpoint_architecture_detail}
shows the tensor dimensions and decoding path.

\section{Local Internal Coordinates: A Step-by-Step Derivation}
\label{app:local_geometry}

This section explains Equations~\ref{eq:local_circle}--\ref{eq:local_decode}
using right triangles, vector projections, and circle coordinates. The
parents $\mathbf{A},\mathbf{B}$ are already placed. The new residue $\mathbf{M}$ is a midpoint in
\emph{sequence index}, not necessarily the spatial midpoint $(\mathbf{A}+\mathbf{B})/2$.
We assume distinct parents and nonnegative radii throughout the derivation.

\subsection{From Two Distances to a Circle}

Fixing $r_L=\|\mathbf{M}-\mathbf{A}\|$ restricts $\mathbf{M}$ to a sphere centered at $\mathbf{A}$.
Fixing $r_R=\|\mathbf{M}-\mathbf{B}\|$ restricts it to a second sphere centered at $\mathbf{B}$.
Their intersection is generally a circle: the distances determine its
position and radius, but leave the position of $\mathbf{M}$ around that circle free.

Let the parent separation be $d=\|\mathbf{B}-\mathbf{A}\|$ and the unit vector
from $\mathbf{A}$ to $\mathbf{B}$ be
$\mathbf{e}_1=(\mathbf{B}-\mathbf{A})/d$. For any candidate $\mathbf{M}$, write
\begin{equation*}
\mathbf{M}=\mathbf{A}+\eta \mathbf{e}_1+\mathbf{v},\qquad \mathbf{v}\cdot \mathbf{e}_1=0.
\end{equation*}
Here the signed distance along the parent axis is $\eta$, and the remaining
displacement perpendicular to it is $\mathbf{v}$. By the Pythagorean theorem,
\begin{equation*}
r_L^2=\eta^2+\|\mathbf{v}\|^2,\qquad
r_R^2=(d-\eta)^2+\|\mathbf{v}\|^2.
\end{equation*}
Subtracting the second equation from the first eliminates $\|\mathbf{v}\|^2$:
\begin{equation*}
r_L^2-r_R^2=\eta^2-(d-\eta)^2=2d\eta-d^2.
\end{equation*}
Thus every point satisfying both distances has the same axial coordinate,
\begin{equation*}
\eta=\frac{d^2+r_L^2-r_R^2}{2d},\qquad \mathbf{C}=\mathbf{A}+\eta \mathbf{e}_1.
\end{equation*}
Its perpendicular distance from $\mathbf{C}$ is also fixed:
\begin{equation*}
\rho=\|\mathbf{v}\|=\sqrt{r_L^2-\eta^2}.
\end{equation*}
All such points therefore lie on a circle of radius $\rho$, centered at $\mathbf{C}$
in a plane perpendicular to $\mathbf{e}_1$. If $r_L=r_R$, then $\eta=d/2$ and $\mathbf{C}$
is the spatial midpoint of the parents. For unequal radii, $\mathbf{C}$ shifts along
the axis and can lie outside the segment $AB$.

The two spheres intersect only when
$|r_L-r_R|\leq d\leq r_L+r_R$. At equality, $\rho=0$ and the circle
reduces to a point. The clamped square root used in the local-ICS geometry decoder prevents
taking the square root of a negative value; it does not by itself make
incompatible distance constraints feasible.

\subsection{Feasibility Repair for Non-intersecting Spheres}
\label{app:sphere_repair}

The main configuration enables geometric repair for non-final midpoint
queries. Let the small interior margin be $\delta=10^{-3}$. If the two spheres
are externally separated, $d>r_L+r_R$, we set
\begin{equation*}
\Delta_{\mathrm{sep}}=
\frac{d+\delta-r_L-r_R}{2},\qquad
(\widetilde r_L,\widetilde r_R)=
(r_L+\Delta_{\mathrm{sep}},r_R+\Delta_{\mathrm{sep}}).
\end{equation*}
Here the separation correction is $\Delta_{\mathrm{sep}}$, and the repaired
radii are $\widetilde r_L,\widetilde r_R$. This equally shares the minimum
Euclidean correction between the two radii
and moves their sum just inside the feasible region. If the left sphere
contains the right one, $r_L-r_R>d$, we instead use
\begin{equation*}
\Delta_{\mathrm{con}}=
\frac{r_L-r_R-\max(d-\delta,0)}{2},\qquad
(\widetilde r_L,\widetilde r_R)=
(r_L-\Delta_{\mathrm{con}},r_R+\Delta_{\mathrm{con}}),
\end{equation*}
where the containment correction is $\Delta_{\mathrm{con}}$. We apply the
symmetric update when $r_R-r_L>d$. This transfers half of the
excess from the larger radius to the smaller one until their difference lies
just inside the containment boundary. Parent coordinates are never moved.

An adjacent-distance constraint takes priority over the symmetric repair. If,
for example, $r_L$ is fixed to its sanitized apo value and $r_R$ is learned,
the implementation leaves $r_L$ unchanged and clamps only the free radius:
\begin{equation*}
\widetilde r_L=r_L,\qquad
\widetilde r_R=
\operatorname{clip}\!\left(
 r_R,\ |d-r_L|+\delta,\ d+r_L-\delta
\right),
\end{equation*}
with the roles reversed when only $r_R$ is fixed. The repaired radii define
$\eta$, $\mathbf{C}$, and $\rho$ in Equation~\ref{eq:local_circle}; the angular state is
normalized before Equation~\ref{eq:local_decode} is applied. We retain the
raw feasibility flag and record whether a projection was applied and its
magnitude. The repair does not re-impose the learned contour-length caps, so a
large correction remains visible as evidence of an upstream parent--radius
mismatch.

Final leaves are handled separately. Both radii are apo-fixed and are never
altered by the non-final repair. If predicted parents make these two
constraints incompatible, the local-ICS geometry decoder evaluates
$\rho=\sqrt{\max(r_L^2-\eta^2,0)}$. Hence $\rho=0$ and $\mathbf{M}=\mathbf{C}$ lies on the parent
axis, while the raw query remains labeled infeasible. This is a finite
numerical fallback, not a claim that both incompatible distances have been
satisfied. During training, the raw violations are additionally penalized by
\begin{equation*}
\mathcal L_{\mathrm{feas}}=
\mathbb E\!\left[
 \operatorname{ReLU}(d-r_L-r_R)^2+
 \operatorname{ReLU}(|r_L-r_R|-d)^2
\right].
\end{equation*}
\rev{Training and inference differ in one respect, and every number in this
paper follows the inference setting. The final-leaf axial fallback above is
applied identically in both. The non-final repair, however, is enabled during
training (\texttt{training.project\_infeasible\_non\_final}) and is \emph{off}
at inference unless requested explicitly through
\texttt{evaluation.apply\_non\_final\_feasibility\_projection}, which defaults
to off. All results reported here, in the main text and in the appendices, were
generated with the non-final feasibility projection switched off at inference.
Enabling it changes almost nothing: re-evaluating the same five checkpoints on
the fixed 150-pair test split with the projection on moves the mean direct
C$\alpha$ RMSD by $+0.00004\pm0.00143$~\AA{} (seed-paired), and in no seed does
the absolute change exceed $0.002$~\AA. The queries that require a correction
are $1.882\pm0.486\%$ of all midpoint queries
(Table~\ref{tab:feasibility_rate}), so the repair acts on a small minority of
tokens and its effect on the resulting structures is below the resolution of
this comparison.}

\subsection{From an Apo Direction to a Local Frame}

\paragraph{Choose a reference vector.}
The circle does not yet have a zero-angle direction. We obtain one from the
apo trace by defining
\begin{equation*}
\bar{\mathbf C}_{\mathrm{apo}}=\frac{\mathbf{A}_{\mathrm{apo}}+\mathbf{B}_{\mathrm{apo}}}{2},\qquad
\mathbf{q}_{\mathrm{apo}}=\mathbf{M}_{\mathrm{apo}}-\bar{\mathbf C}_{\mathrm{apo}}.
\end{equation*}
The vector from the apo parents' spatial midpoint toward the apo midpoint
residue is $\mathbf{q}_{\mathrm{apo}}$. The apo parents' spatial midpoint,
$\bar{\mathbf C}_{\mathrm{apo}}$, is not
the circle center $\mathbf{C}$: the former uses apo coordinates, while the latter
uses the current parents and the two prescribed radii.

\paragraph{Remove the component along the parent axis.}
Since $\mathbf{e}_1$ has unit length, the scalar
$\mathbf{q}_{\mathrm{apo}}\cdot \mathbf{e}_1$ is the signed length of the component along
the axis. Multiplying this scalar by $\mathbf{e}_1$ gives the axial vector itself.
Subtracting it leaves only the component parallel to the circle plane:
\begin{equation*}
\mathbf{q}_\parallel=(\mathbf{q}_{\mathrm{apo}}\cdot \mathbf{e}_1)\mathbf{e}_1,\qquad
\mathbf{q}_\perp=\mathbf{q}_{\mathrm{apo}}-\mathbf{q}_\parallel.
\end{equation*}
Indeed,
$\mathbf{e}_1\cdot \mathbf{q}_\perp=\mathbf{e}_1\cdot \mathbf{q}_{\mathrm{apo}}
-(\mathbf{q}_{\mathrm{apo}}\cdot \mathbf{e}_1)(\mathbf{e}_1\cdot \mathbf{e}_1)=0$.
In matrix notation, the same operation is
\begin{equation*}
\mathbf{q}_\perp=(\mathbf{I}-\mathbf{e}_1\mathbf{e}_1^\top)\mathbf{q}_{\mathrm{apo}},
\end{equation*}
where $\mathbf{I}$ is the $3\times3$ identity matrix. The outer product $\mathbf{e}_1\mathbf{e}_1^\top$
extracts the axial component; subtracting it from $\mathbf{I}$ keeps the perpendicular
component.

\paragraph{Why translation followed by projection is the same construction.}
The target plane passes through $\mathbf{C}$ and is \emph{perpendicular} to $\mathbf{e}_1$;
it is not a plane containing the parent axis. Denote this circle plane by
$\Pi_{\mathbf C}$:
\begin{equation*}
\Pi_{\mathbf C}=\{\mathbf{P}:(\mathbf{P}-\mathbf{C})\cdot \mathbf{e}_1=0\}.
\end{equation*}
Translate the arrow representing $\mathbf{q}_{\mathrm{apo}}$ so that its tail moves
from $\bar{\mathbf C}_{\mathrm{apo}}$ to $\mathbf{C}$. Its length and direction do not change.
Its translated head is $\mathbf{T}=\mathbf{C}+\mathbf{q}_{\mathrm{apo}}$.
The tail already lies in $\Pi_{\mathbf C}$.
To project the head onto that plane, subtract its signed normal displacement:
\begin{align*}
\mathbf{T}_{\mathrm{proj}}
&=\mathbf{T}-\bigl[(\mathbf{T}-\mathbf{C})\cdot \mathbf{e}_1\bigr]\mathbf{e}_1\\
&=\mathbf{C}+\mathbf{q}_{\mathrm{apo}}-(\mathbf{q}_{\mathrm{apo}}\cdot \mathbf{e}_1)\mathbf{e}_1\\
&=\mathbf{C}+\mathbf{q}_\perp.
\end{align*}
The projected arrow from $\mathbf{C}$ to $\mathbf{T}_{\mathrm{proj}}$ is therefore exactly
$\mathbf{q}_\perp$. Translating the arrow chooses its location; removing the axial
component determines its projection. No rotation of the apo vector is
involved. Both constructions use the axis of the \emph{current} parents,
not the apo parent axis.

\paragraph{Normalize the direction and complete the frame.}
The projected arrow need not reach the circle. Only its direction is needed:
\begin{equation*}
\mathbf{e}_2=\frac{\mathbf{q}_\perp}{\|\mathbf{q}_\perp\|},\qquad \mathbf{e}_3=\mathbf{e}_1\times \mathbf{e}_2.
\end{equation*}
The vectors $\mathbf{e}_2,\mathbf{e}_3$ are perpendicular unit vectors in the circle plane;
the cross product fixes their orientation by the right-hand rule. The
zero-angle point is $\mathbf{C}+\rho \mathbf{e}_2$, not necessarily $\mathbf{M}_{\mathrm{apo}}$ or
$\mathbf{T}_{\mathrm{proj}}$. This normalization requires a nonzero projected vector;
the degenerate case is discussed next.

\subsection{Degenerate References and the Axis Fallback}
\label{app:frame_fallback}

\paragraph{Why a fallback is needed.}
If $\mathbf{q}_{\mathrm{apo}}$ is parallel to $\mathbf{e}_1$, removing its axial component
leaves $\mathbf{q}_\perp=0$. The same happens if $\mathbf{q}_{\mathrm{apo}}$ itself is zero.
There is then no apo-defined direction in the circle plane, and dividing by
$\|\mathbf{q}_\perp\|$ is undefined. A near-zero projection also makes the normalized
direction sensitive to small numerical perturbations. The implementation
activates the fallback when the projected norm is at most $10^{-6}$ during
target encoding or at most $10^{-8}$ during Cartesian decoding. The fallback
supplies a reference direction for these cases; it does not change the circle
center or the prescribed radii.

\paragraph{Least-aligned coordinate-axis construction.}
For a fallback based on the fixed Cartesian unit axes
$\mathbf{a}_x=(1,0,0)$, $\mathbf{a}_y=(0,1,0)$, and $\mathbf{a}_z=(0,0,1)$, the selection rule is
\begin{equation*}
\mathbf{a}_*\in\underset{\mathbf{a}\in\{\mathbf{a}_x,\mathbf{a}_y,\mathbf{a}_z\}}{\arg\min}|\mathbf{a}\cdot \mathbf{e}_1|.
\end{equation*}
The absolute dot product measures how closely an axis aligns with $\mathbf{e}_1$,
regardless of sign. Choosing the smallest value selects the axis closest to
being perpendicular to $\mathbf{e}_1$. The implementation tests the axes in
$x,y,z$ order, and the first minimum is selected in a tie. The replacement
frame is then
\begin{equation*}
\mathbf{v}_{\mathrm{fallback}}=\mathbf{a}_*-(\mathbf{a}_*\cdot \mathbf{e}_1)\mathbf{e}_1,\qquad
\mathbf{e}_2=\frac{\mathbf{v}_{\mathrm{fallback}}}{\|\mathbf{v}_{\mathrm{fallback}}\|},\qquad
\mathbf{e}_3=\mathbf{e}_1\times \mathbf{e}_2.
\end{equation*}
This projection cannot vanish for a unit $\mathbf{e}_1$. Its three squared Cartesian
components sum to one, so the smallest is at most $1/3$. Consequently,
\begin{equation*}
\|\mathbf{v}_{\mathrm{fallback}}\|^2=1-(\mathbf{a}_*\cdot \mathbf{e}_1)^2\geq\frac{2}{3}.
\end{equation*}
For example, when $\mathbf{e}_1$ points along the $x$-axis, either the $y$- or
$z$-axis provides a perpendicular reference direction.

\paragraph{Determinism is not rotation equivariance.}
For the nondegenerate apo-based frame, rotating both structures rotates
$\mathbf{e}_1$ and $\mathbf{q}_{\mathrm{apo}}$ together, so their projection rotates with
them. Fixed Cartesian axes do not rotate with the protein. A fallback chosen
from those axes is reproducible in a given coordinate frame, but need not
produce the rotated version of the original $\mathbf{e}_2$ after the protein is
rotated. It therefore does not, in general, preserve rotation equivariance
of the local decoder when activated. This limitation is distinct from the
equivariance of the Cartesian anchor network in Section~\ref{sec:top_egnn}.

Coincident parents ($d=0$) are a separate degeneracy: the parent axis itself
is undefined, so replacing the apo reference vector alone does not resolve
that case. At sphere tangency ($\rho=0$), the decoded point is simply $\mathbf{C}$,
independent of the angular reference.

\subsection{Placing the Midpoint: A Worked Example}

In a two-dimensional plane, a circle of radius $\rho$ has coordinates
$(\rho\cos\omega,\rho\sin\omega)$. Replacing the two coordinate axes by
$\mathbf{e}_2,\mathbf{e}_3$ and placing the origin at $\mathbf{C}$ gives
\begin{equation*}
\mathbf{M}=\mathbf{C}+\rho(\cos\omega\,\mathbf{e}_2+\sin\omega\,\mathbf{e}_3).
\end{equation*}
At $\omega=0$, the displacement points along $+\mathbf{e}_2$; at
$\omega=\pi/2$, it points along $+\mathbf{e}_3$. Positive angles thus run from
$+\mathbf{e}_2$ toward $+\mathbf{e}_3$. Changing $\omega$ leaves both parent distances
unchanged because $\cos^2\omega+\sin^2\omega=1$.

For a simple numerical example, let $\mathbf{A}=(0,0,0)$, $\mathbf{B}=(6,0,0)$, and
$r_L=r_R=5$, in arbitrary length units. Then
\begin{equation*}
d=6,\quad \mathbf{e}_1=(1,0,0),\quad \eta=3,\quad
\mathbf{C}=(3,0,0),\quad \rho=\sqrt{25-9}=4.
\end{equation*}
The circle lies in the plane $x=3$. Suppose the apo reference vector is
$\mathbf{q}_{\mathrm{apo}}=(2,3,4)$. Moving its tail to $\mathbf{C}$ places its head at
$\mathbf{T}=(5,3,4)$. Orthogonal projection onto $x=3$ gives
$\mathbf{T}_{\mathrm{proj}}=(3,3,4)$, so the projected arrow is $(0,3,4)$.
Subtracting the axial component gives the same result directly:
\begin{equation*}
\mathbf{q}_\perp=(2,3,4)-2(1,0,0)=(0,3,4).
\end{equation*}
Normalizing and taking the cross product yields
$\mathbf{e}_2=(0,3/5,4/5)$ and $\mathbf{e}_3=(0,-4/5,3/5)$. At $\omega=0$,
\begin{equation*}
\mathbf{M}=\mathbf{C}+4\mathbf{e}_2=(3,12/5,16/5).
\end{equation*}
Both squared parent distances are
$3^2+(12/5)^2+(16/5)^2=25$, as required. Notice that the projected apo
arrow has length $5$, while the circle radius is $4$: projection determines
the reference direction, and the two parent distances determine the radius.

\subsection{Empirical Support for the Adjacent-Distance Prior}
\label{app:adjacent_distance_stats}

We characterize the apo-distance assumption over all 1,591 D3PM pairs as a
post-hoc dataset audit; these statistics were not used to select a model or a
hyperparameter. Segment boundaries are essential: treating consecutive rows
of a compressed tensor as bonded residues would introduce false adjacencies
across gaps. We restore the curated residue masks for 1,587 pairs and recover
the four remaining 6XR4 pairs from their source PDB files. This yields 577,291
within-segment consecutive C$\alpha$ pairs and excludes 3,960 transitions
across mask-defined gaps.

\begin{figure}[htbp]
\centering
\includegraphics[width=\linewidth]{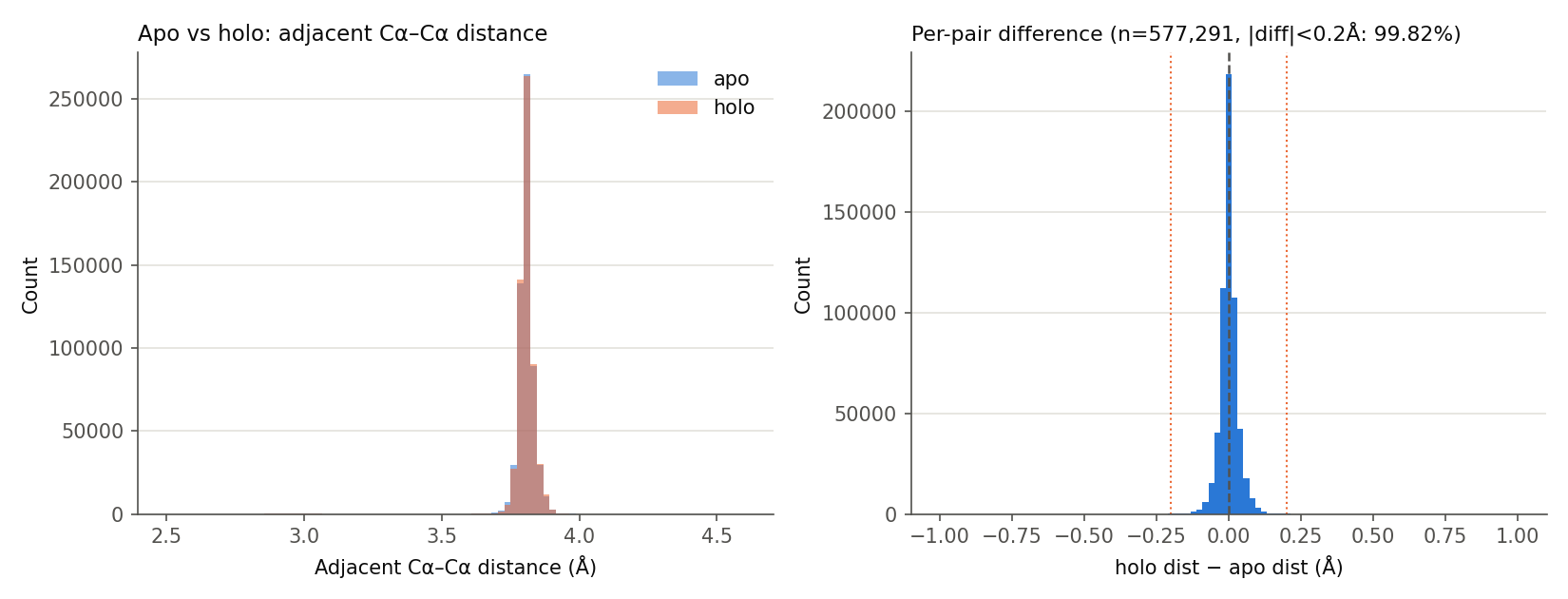}
\caption{Adjacent C$\alpha$--C$\alpha$ distances in matched apo and holo
structures. The left panel overlays the two distance distributions. The right
panel shows $d_{\mathrm{holo}}-d_{\mathrm{apo}}$; orange dotted lines mark
$\pm0.2$~\AA. The analysis contains 577,291 within-segment consecutive
C$\alpha$ pairs from all 1,591 D3PM pairs, of which 99.8202\% lie within the
marked interval.}
\label{fig:apo_holo_adjacent}
\end{figure}

Among these pairs, 576,253/577,291 (99.8202\%) satisfy
$|d_{\mathrm{holo}}-d_{\mathrm{apo}}|<0.2$~\AA. The mean and median absolute
differences are 0.0226 and 0.0147~\AA, respectively. Figure~\ref{fig:apo_holo_adjacent}
shows the full adjacent-distance distributions and their signed differences.

\subsection{Test-Time Sphere-Feasibility Diagnostics}
\label{app:feasibility_rate}

Table~\ref{tab:feasibility_rate} reports raw sphere-intersection violations
during free recursive rollout on the fixed 150-pair test split. The
\texttt{feasible} flag is evaluated from the predicted radii before geometric
repair. A non-final query can therefore be counted here even though its radii
are subsequently projected before the midpoint is placed.

\begin{table}[htbp]
\centering
\small
\setlength{\tabcolsep}{5pt}
\begin{tabular}{rrrrr}
\toprule
Training seed & Midpoint queries & Raw infeasible & Rate (\%)
& \shortstack{Non-final / final-leaf\\events}\\
\midrule
1  & 40,724 & 564 & 1.3849 & 532 / 32\\
3  & 40,724 & 685 & 1.6821 & 646 / 39\\
42 & 40,724 & 996 & 2.4457 & 972 / 24\\
88 & 40,724 & 960 & 2.3573 & 940 / 20\\
99 & 40,724 & 628 & 1.5421 & 606 / 22\\
\midrule
Pooled & 203,620 & 3,833 & 1.8824 & 3,696 / 137\\
\bottomrule
\end{tabular}
\caption{Raw pre-repair sphere infeasibility for the five base models. The
per-seed rate is $1.882\pm0.486\%$ (mean $\pm$ sample standard deviation).
Non-final radii are eligible for the repair in Appendix~\ref{app:sphere_repair};
final leaves retain both apo-fixed radii and use the finite axial fallback if
needed.}
\label{tab:feasibility_rate}
\end{table}

Of the 3,833 raw violations, 3,696 (96.43\%) are non-final queries and 137
(3.57\%) are final leaves. Final-leaf violations constitute 0.0673\% of all
generated midpoint queries. The saved evaluation artifacts do not contain the
total number of final-leaf opportunities, so we do not report a
category-conditional final-leaf rate.

\section{Main Configuration}
\label{app:implementation}

\begin{table}[ht]
\caption{Configuration used for the five-seed main result.}
\label{tab:implementation}
\centering
\small
\begin{tabular}{ll}
\toprule
Component & Setting\\
\midrule
Data split & 1,291 / 150 / 150 apo--holo pairs\\
Residue features & D3PM 145 + precomputed ESM-3 1,536\\
Top anchor stride & 4\\
Top model & EGNN flow, 8 layers, node 320, edge bottleneck 160\\
Top graph & apo 32-NN $\cup$ initial-state 16-NN $\cup$ sequence edges\\
Top RBF / sequence features & 32 current + 32 apo + 17 Fourier channels\\
Global node context & masked mean over valid anchors\\
Midpoint model & Transformer, 8 layers, 8 heads, token 256, condition 128\\
Midpoint registers & 2\\
Parent feature & normalized distance + 16 Gaussian RBF channels\\
Midpoint geometry & local internal coordinates with 3/2/1-DOF policy\\
Flow parameterization & clean endpoint prediction\\
Time sampling & $t=0$ / Euler grid / uniform = 1:1:1\\
Inference & 5 Euler steps per active flow\\
Training & AdamW, batch 8, 1,000 epochs\\
Checkpoint selection & minimum validation mean Kabsch RMSD\\
Training seeds & 1, 3, 42, 88, 99\\
\bottomrule
\end{tabular}
\end{table}

The top distance RBF centers span 0--6 normalized coordinate units; this is
not a 6~\AA{} graph cutoff. The midpoint parent-distance RBF centers span
0--64~\AA. The top graph indices are fixed within a forward pass, but current
edge distances are recomputed after each coordinate update. Self-conditioning,
scheduled sampling, and a separate previous-level memory module are disabled
in the main configuration.

Coordinate normalizers are estimated from training-set holo top anchors:
$\widetilde{\mathbf X}=(\mathbf{X}-\boldsymbol{\mu}_{\mathbf X})/\boldsymbol{\sigma}_{\mathbf X}$, with per-channel population mean and
standard deviation. Midpoint states use the analogous channel-wise statistics
over all training targets, $\widetilde{\mathbf S}=(\mathbf{S}-\boldsymbol{\mu}_{\mathbf S})/\boldsymbol{\sigma}_{\mathbf S}$. Predictions are
mapped back with the inverse transformations before geometry decoding and
evaluation. The midpoint parent feature concatenates $d/64$ with 16 Gaussian
RBF values whose centers are uniformly spaced from 0 to 64~\AA.

\section{Training Objective Details}
\label{app:loss_details}

The main configuration predicts clean endpoints. Let the normalized holo
anchor coordinates be $\widetilde{\mathbf X}_1$, and let the standardized
four-channel local state for query $q$ at midpoint level $\ell$ be
$\widetilde{\mathbf S}_{1,q}^{(\ell)}$. The two primary losses are
\begin{align}
\mathcal L_{\mathrm{top}}
&=\frac{1}{3|\mathcal A|}\sum_{i\in\mathcal A}
  \left\|f_{\theta}^{\mathrm{top}}(\mathbf{X}_t,t,\mathbf{F})_i-
  \widetilde{\mathbf X}_{1,i}\right\|_2^2,\\
\mathcal L_{\mathrm{mid}}
&=\frac{1}{D}\sum_{\ell=1}^{D}\frac{1}{Q_\ell}
  \sum_{q=1}^{Q_\ell}\left[
  \frac{1}{2}\sum_{c=1}^{2}a_{q,c}e_{q,c}^2+
  \frac{1}{2}\sum_{c=3}^{4}e_{q,c}^2\right],
\label{eq:loss_primary_details}
\end{align}
where the number of active midpoint queries at level $\ell$ is $Q_\ell$, and
the clean-state prediction error for query $q$ and channel $c$ is $e_{q,c}$.
The first two midpoint channels
encode radii and the last two encode $(\sin\omega,\cos\omega)$. The mask
$a_{q,c}$ is zero when Equation~\ref{eq:adjacent_constraint} fixes that radius
and one otherwise. Thus the model receives no state-loss gradient for a radius
that the decoder replaces by an apo distance. The loss is averaged over all
queries within each level and equally over levels; apo-fixed radius channels
contribute zero but retain the two-channel normalization.

For the Cartesian auxiliary terms, define the Smooth-L1 penalty
\begin{equation}
h_\beta(e)=
\begin{cases}
e^2/(2\beta), & |e|<\beta,\\
|e|-\beta/2, & \text{otherwise}.
\end{cases}
\label{eq:smooth_l1}
\end{equation}
The geometric auxiliary losses below are evaluated on the teacher-forced
partial reconstruction used during training. The decoded-coordinate loss
$\mathcal L_{\mathrm{xyz}}$ averages
$h_{0.1}(\widehat{\mathbf M}-\mathbf{M}_{\mathrm{holo}})$ over Cartesian components, queries,
and midpoint levels. For a contiguous segment $g$ and sequence offset
$\Delta$, let
\begin{equation}
\ell_{\Delta,g}(\widehat{\mathbf X},\mathbf{X}_1)=
\frac{1}{|\mathcal P_{\Delta,g}|}
\sum_{i\in\mathcal P_{\Delta,g}}
h_\beta\!\left(
\|\widehat{\mathbf X}_{i+\Delta}-\widehat{\mathbf X}_i\|-
\|\mathbf{X}_{1,i+\Delta}-\mathbf{X}_{1,i}\|
\right),
\end{equation}
where the set of valid index pairs separated by $\Delta$ within segment $g$ is
$\mathcal P_{\Delta,g}$. Then
$\mathcal L_{\mathrm{adj}}$ is the mean of $\ell_{1,g}$ over segments with
$\beta=0.05$. The sparse multiscale term is
\begin{equation}
\mathcal L_{\mathrm{sparse}}=
\frac{\sum_{\Delta\in\{1,2,4\}}\sum_g
w_\Delta\ell_{\Delta,g}}
{\sum_{\Delta\in\{1,2,4\}}\sum_g w_\Delta},
\qquad (w_1,w_2,w_4)=(1,1,0.5),
\label{eq:sparse_loss_details}
\end{equation}
using $\beta=0.1$. Offsets one, two, and four supervise connectivity, local
bending, and slightly wider local shape without constructing all $L^2$ pairs.

Before any non-final repair, let the parent separation and two decoded radii
for query $q$ be $d_q$, $r_{L,q}$, and $r_{R,q}$, respectively. The feasibility
penalty is averaged within each midpoint level and then equally across levels:
\begin{equation}
\mathcal L_{\mathrm{feas}}=
\frac{1}{D}\sum_{\ell=1}^{D}\frac{1}{Q_\ell}
\sum_{q=1}^{Q_\ell}\!\left[
\operatorname{ReLU}(d_q-r_{L,q}-r_{R,q})^2+
\operatorname{ReLU}(|r_{L,q}-r_{R,q}|-d_q)^2
\right].
\label{eq:feasibility_loss_details}
\end{equation}
The full implemented objective is
\begin{equation}
\mathcal L=\mathcal L_{\mathrm{top}}+\mathcal L_{\mathrm{mid}}
+\alpha_s\left(
0.2\mathcal L_{\mathrm{xyz}}+0.1\mathcal L_{\mathrm{adj}}
+0.1\mathcal L_{\mathrm{sparse}}+0.05\mathcal L_{\mathrm{feas}}
\right),
\end{equation}
with a linear auxiliary warm-up
$\alpha_s=\min(\max(s/500,0),1)$ at optimization step $s$. Optional
top-coordinate, pairwise, endpoint, and RMSD auxiliaries are implemented but
have zero weight in the reported configuration.

\section{Model Size and Inference Cost}
\label{app:compute}

\begin{table}[htbp]
\centering
\scriptsize
\setlength{\tabcolsep}{3.2pt}
\resizebox{\linewidth}{!}{%
\begin{tabular}{lrrrrr}
\toprule
Method & Parameters & Solver steps & Hardware & Test wall time & Time/target\\
\midrule
\method{} (ours) & 34.080M & 5 per active stage & RTX 2080 Ti 11GB
& 36.20 s & 0.241 s\\
Direct-regression EGNN (local control) & 0.165M & 1 & RTX 2080 Ti 11GB
& 2.30 s & 0.015 s\\
SBAlign & 0.545M & 100 & RTX 2080 Ti 11GB & 245.86 s & 1.639 s\\
ABM (FDBM implementation) & 0.538M & 100 & RTX 2080 Ti 11GB & 200.86 s & 1.339 s\\
FDBM & 0.538M & 100 & RTX 2080 Ti 11GB & 208.24 s & 1.388 s\\
\bottomrule
\end{tabular}%
}
\caption{Hardware-controlled inference cost on the same 150-pair D3PM test
split, with one generated structure per target. Every method was run
sequentially on one NVIDIA GeForce RTX 2080 Ti. Wall time is measured around
the CUDA-synchronized inference loop and includes per-target data transfer,
sampling, and C$\alpha$ metric computation; it excludes Python startup,
checkpoint and dataset initialization, prediction serialization, ESM-3
preprocessing, cg2all reconstruction, and MolProbity.}
\label{tab:compute}
\end{table}

\method{} has 34,080,066 registered trainable parameters: 23,562,666 in the
top-anchor EGNN and 10,517,400 in the midpoint Transformer. The count excludes
ESM-3 because embeddings are precomputed. In the hardware-controlled run,
\method{}, direct-regression EGNN, SBAlign, ABM, and FDBM take 36.20, 2.30,
245.86, 200.86, and 208.24 seconds, respectively, for the complete 150-target
test inference loop. These are single-checkpoint runtime measurements rather
than seed-averaged accuracy estimates. \method{} performs five Euler
updates for the top-anchor stage and five updates at each active recursive
midpoint level, whereas the bridge baselines use 100 integration steps and the
direct EGNN uses one forward pass. The local direct-EGNN control is distinct
from the legacy \texttt{EGNN (reported)} row in Table~\ref{tab:accuracy}.

\section{Full-Sequence ESM-3 Mapping}
\label{app:esm_mapping}

We map each canonical D3PM row first to an ordered apo ATOM residue and then to
the deposited polymer sequence. Author residue numbers and insertion codes are
used when available. If insertion codes were lost during dataset construction,
positions with the same residue number are disambiguated by amino-acid
identity. Four pairs require a global-alignment fallback. All 1,591 pairs map
without changing a deposited amino acid. Eight pairs contain a non-monotone
gather order because of insertion-code ordering or circular permutation.

We first run ESM-3 on the entire deposited apo polymer sequence (SEQRES),
including residues that are unresolved in the apo coordinates. For sequences
of at most 600 residues, ESM-3 is applied once. Longer sequences use length-600
windows with stride 300. If a residue occurs in several windows, we retain the
representation from the window in which it is farthest from a boundary. BOS
and EOS representations are averaged across windows. We then gather only the
embeddings corresponding to observed D3PM residues, in D3PM graph order.

\section{Rotation, Translation, and Reflection Consistency}
\label{app:equivariance}

An equivariant model should give the same structure when the input protein is
shown in a different position or orientation. The output should only move with
the input. We test this property on all 150 test pairs using each of the five
trained models. First, we generate a prediction from the original input. We
then transform every Cartesian input and run the model again. We use the same
Gaussian midpoint-noise seed in both runs, pairing the stochastic draws
channel-wise. Finally, we apply the inverse transform to the second prediction.
If the model is equivariant to that transform, the two predictions should
coincide without structural alignment.

We test rotation, translation, rotation plus translation, and reflection. A
proper rotation is obtained by QR-orthogonalizing a $3\times3$ Gaussian matrix,
correcting its column signs, and enforcing determinant $+1$. For reflection,
we independently sample a proper rotation and flip its last column, producing
a Haar-random improper orthogonal matrix $\mathbf{Q}$ with $\mathbf{Q}^\top \mathbf{Q}=\mathbf{I}$ and
$\det \mathbf{Q}=-1$. We draw one transform per test pair with a fixed random seed.
Each translation component is sampled from $[-25,25]$~\AA. All coordinate
fields, including the apo reference vectors used by the local decoder, receive
the same transform.

The two RMSD columns answer different questions. The \emph{full-trace
difference} directly compares the original prediction with the
inverse-transformed new prediction, without Kabsch alignment. It is the strict
equivariance check. The \emph{change in holo RMSD} first computes the
Kabsch-aligned RMSD of each prediction to its corresponding holo target and
then reports the absolute difference between those two errors. It checks
whether prediction accuracy is preserved, but does not by itself prove that the
two generated structures are identical.

\begin{table}[htbp]
\centering
\scriptsize
\setlength{\tabcolsep}{3.2pt}
\resizebox{\linewidth}{!}{%
\begin{tabular}{lrrrr}
\toprule
Transform
& \shortstack{Top-anchor\\difference (\AA)}
& \shortstack{Full-trace\\difference (\AA)}
& \shortstack{Change in\\holo RMSD (\AA)}
& \shortstack{Max. full-trace\\difference (\AA)}\\
\midrule
Rotation
& $3.245{\times}10^{-4}\!\pm\!1.9{\times}10^{-5}$
& $4.998{\times}10^{-4}\!\pm\!2.5{\times}10^{-5}$
& $3.708{\times}10^{-5}\!\pm\!3.7{\times}10^{-6}$
& $5.375{\times}10^{-3}$\\
Translation
& $3.231{\times}10^{-4}\!\pm\!3.2{\times}10^{-5}$
& $4.872{\times}10^{-4}\!\pm\!3.6{\times}10^{-5}$
& $3.811{\times}10^{-5}\!\pm\!4.2{\times}10^{-6}$
& $5.788{\times}10^{-3}$\\
Rotation $+$ translation
& $3.252{\times}10^{-4}\!\pm\!2.1{\times}10^{-5}$
& $4.904{\times}10^{-4}\!\pm\!3.7{\times}10^{-5}$
& $3.625{\times}10^{-5}\!\pm\!5.5{\times}10^{-6}$
& $5.554{\times}10^{-3}$\\
Reflection
& $2.576{\times}10^{-4}\!\pm\!1.4{\times}10^{-4}$
& $1.633\!\pm\!0.027$
& $0.725\!\pm\!0.014$
& $4.442$\\
\bottomrule
\end{tabular}%
}
\caption{Consistency under rigid transforms and reflection on the fixed
150-pair test split. We invert each input transform before comparing the new
prediction with the original prediction; no Kabsch alignment is used. Values
are means $\pm$ sample standard deviations across five trained models. The last
column is the largest difference among all $5\times150$ predictions.}
\label{tab:equivariance}
\end{table}

For proper rotations and translations, the inverse-transformed full traces
differ by about $0.0005$~\AA{} on average, and the largest difference is below
$0.0058$~\AA. Holo RMSD changes by less than $0.000039$~\AA{} on average. Thus,
the composed generator is numerically consistent under the SE(3) transforms
that preserve molecular handedness.

Reflection gives a different result. The top-anchor difference remains only
$(2.576\pm1.4)\times10^{-4}$~\AA, confirming numerical reflection equivariance
of the Cartesian EGNN. However, the complete trace differs by
$1.633\pm0.027$~\AA, and holo RMSD changes by $0.725\pm0.014$~\AA. The adjacent-
distance MAE changes by only $(6.49\pm3.37)\times10^{-7}$~\AA{} because the
radial constraints remain distance based, whereas pseudo-angle MAE changes by
$3.915\pm0.024$ degrees. The composed model is therefore not reflection-
equivariant. This behavior arises after the anchor stage: reflection reverses
handedness, changes the parity of the cross-product axis
$\mathbf{e}_3=\mathbf{e}_1\times \mathbf{e}_2$, and is not an enforced symmetry of the midpoint
Transformer.

At the C$\alpha$-trace level, reflection produces the mirror geometry associated
with the opposite handedness rather than another pose of the same L-protein.
L- and D-proteins are biologically distinct because chirality changes their
stereochemistry and molecular interactions. Reflection therefore lies outside
the intended SE(3) symmetry group for L-protein conformations. We require
consistency under proper rotations and translations, which preserve
handedness. This fixed-seed experiment is a samplewise check, not
a test of distribution-level symmetry under alternative latent-noise couplings.
It is also empirical rather than a proof for every input. In addition, the
degenerate-frame fallback in Appendix~\ref{app:frame_fallback} uses a
fixed global axis and remains an exception to exact rotation equivariance.

\section{Detailed Structural Validity}
\label{app:physical_details}

Main Table~\ref{tab:physical} retains the three summary measures used in
Figure~\ref{fig:molprobity_benchmark}. Table~\ref{tab:physical_detailed} reports
the remaining C$\alpha$ and all-atom diagnostics. Our values are mean $\pm$
sample standard deviation over five independently trained models.

\begin{table}[H]
\caption{Detailed C$\alpha$ and relaxation-free cg2all--MolProbity metrics. C$\alpha$ clashes count non-bonded pairs; only direct sequence neighbours within the same restored segment are excluded.}
\label{tab:physical_detailed}
\centering
\scriptsize
\setlength{\tabcolsep}{3.3pt}
\resizebox{\linewidth}{!}{%
\begin{tabular}{lrrrrrr}
\toprule
Method & C$\alpha$ clash/$10^6$ $\downarrow$ & Clashscore $\downarrow$
& Rama fav. (\%) $\uparrow$ & Rama out. (\%) $\downarrow$
& Bond RMSD (\AA) $\downarrow$ & Angle RMSD (deg) $\downarrow$\\
\midrule
GT-holo C$\alpha$ control & -- & 38.3 & 96.83 & 0.35 & 0.0205 & 2.00\\
Direct-regression EGNN & 2794.18 & 507.7 & 26.65 & 51.05 & 0.1773 & 14.48\\
SBAlign & 270.95 & 415.8 & 35.78 & 39.98 & 0.1680 & 11.69\\
FDBM & 113.09 & 261.7 & 63.77 & 18.72 & 0.1014 & 9.15\\
ABM & 121.43 & 264.6 & 63.15 & 18.97 & 0.1019 & 9.30\\
\textbf{\method{} (ours)}
& $\mathbf{20.01\!\pm\!3.27}$
& $\mathbf{95.35\!\pm\!6.64}$
& $\mathbf{90.14\!\pm\!0.45}$
& $\mathbf{3.12\!\pm\!0.26}$
& $\mathbf{0.0242\!\pm\!0.0006}$
& $\mathbf{2.817\!\pm\!0.071}$\\
\bottomrule
\end{tabular}
}
\end{table}

\section{Validation-Based Hyperparameter Selection}
\label{app:ablation}

\begin{figure}[t]
\centering
\includegraphics[width=\linewidth]{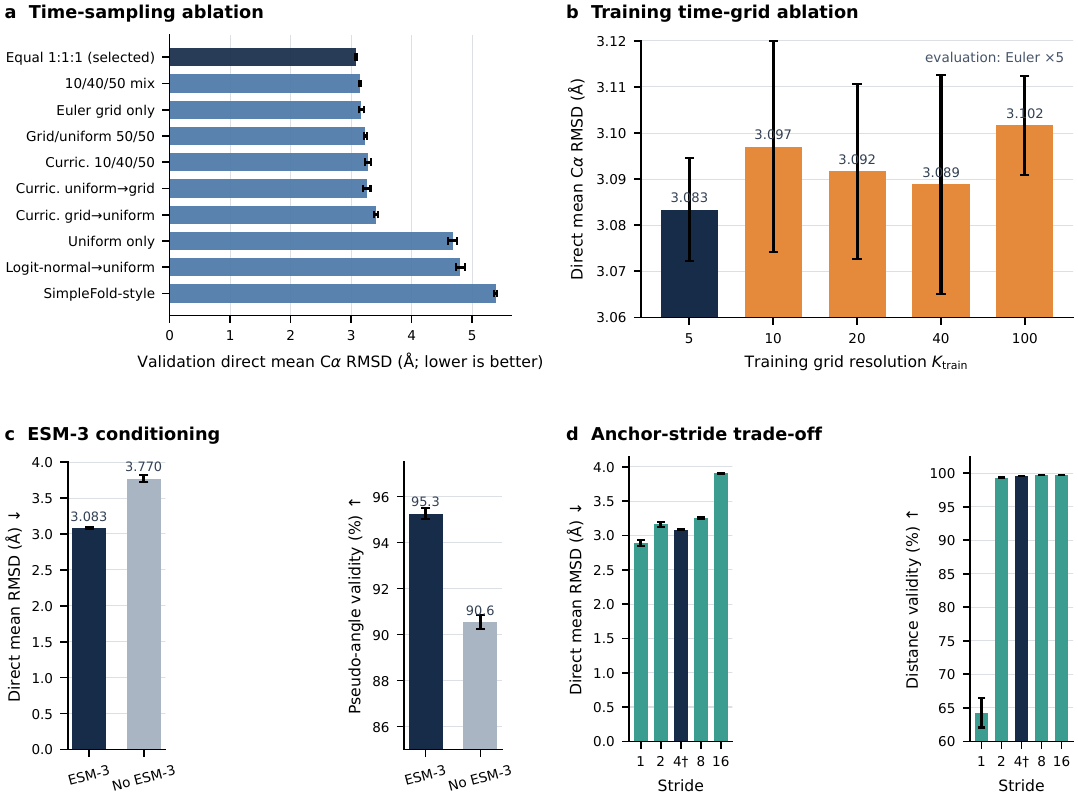}
\caption{Validation ablations used for model selection. Bars and error bars
show means and sample standard deviations across five independently trained
models. (a) Absolute direct mean RMSD for every time sampler, including the selected
equal 1:1:1 mixture of $t=0$, Euler-grid, and uniform samples. (b) Training-grid
resolution; evaluation uses five Euler steps in every condition. (c) ESM-3
conditioning improves direct RMSD and pseudo-angle fidelity. (d) Stride 4 balances
direct RMSD and adjacent-distance fidelity. Dark bars and daggers denote the
selected configuration.}
\label{fig:validation_ablation}
\end{figure}

We selected time sampling, training-grid resolution, ESM-3 conditioning, and
anchor stride on the validation split. \rev{The original selection matrix
contains 19 conditions and five training seeds per condition, for 95 trained
models in total. Additional post-selection mechanism controls are reported
separately in Appendix~\ref{app:mechanistic_controls} and were not used to
revise the test-set results.} Every model is evaluated on the same 150 validation pairs. Fixed-C$\alpha$
cg2all reconstruction and MolProbity use the same 149 eligible pairs per run;
the excluded pair, \texttt{4P7G\_C\_3OE4\_A}, consists of four disconnected
singleton segments and does not define a local frame for backmapping.
The RMSD columns use the public-protocol direct metric in the pre-aligned D3PM
frame. Checkpoints were nevertheless selected by the prespecified validation
Kabsch RMSD. Local metrics pool the same 55,064 within-segment adjacent pairs
and 54,571 contiguous triplets for every run.
Table~\ref{tab:validation_ablation} reports representative global, local, and
all-atom metrics. Figure~\ref{fig:validation_ablation} visualizes the
factor-wise comparisons and the accuracy--validity trade-off. The held-out
test results are not used in this analysis.

\begin{table}[htbp]
\caption{Validation ablations used for model selection. Values are mean $\pm$ sample SD over five training seeds. The selected base (marked $\dagger$) uses equal-probability $t=0$/Euler-grid/uniform sampling, $K_{\mathrm{train}}=5$, stride 4, and ESM-3. Distance and angle validity denote errors within 0.2~\AA{} and $20^\circ$, respectively. Direct RMSD is measured in the shared pre-aligned D3PM frame without a post-hoc prediction-to-target Kabsch fit. All conditions use five Euler integration steps at evaluation.}
\label{tab:validation_ablation}
\centering
\scriptsize
\setlength{\tabcolsep}{3.0pt}
\resizebox{\linewidth}{!}{%
\begin{tabular}{@{}lrrrrr@{}}
\toprule
Setting & Direct mean RMSD (\AA{}) $\downarrow$ & Distance valid (\%) $\uparrow$ & Angle valid (\%) $\uparrow$ & Clash/$10^6$ $\downarrow$ & MP score $\downarrow$\\
\midrule
\textbf{Selected base}$^\dagger$ & $3.083\pm0.011$ & $99.632\pm0.021$ & $95.261\pm0.225$ & $25.18\pm3.73$ & $3.044\pm0.064$\\
\addlinespace[2pt]
\multicolumn{6}{@{}l}{\textit{Time sampling}}\\
\quad $t_0$/grid/uniform = 10/40/50 & $3.148\pm0.022$ & $99.618\pm0.033$ & $94.745\pm0.200$ & $27.85\pm3.92$ & $3.133\pm0.020$\\
\quad Euler grid only & $3.165\pm0.042$ & $99.617\pm0.013$ & $94.974\pm0.153$ & $25.37\pm4.78$ & $3.119\pm0.065$\\
\quad Grid/uniform = 50/50 & $3.232\pm0.023$ & $99.627\pm0.033$ & $94.440\pm0.164$ & $27.38\pm1.69$ & $3.202\pm0.029$\\
\quad Curriculum: $t_0\rightarrow$grid$\rightarrow$uniform & $3.280\pm0.051$ & $99.630\pm0.014$ & $94.062\pm0.262$ & $30.14\pm1.66$ & $3.181\pm0.030$\\
\quad Curriculum: uniform$\rightarrow$grid & $3.260\pm0.060$ & $99.593\pm0.046$ & $94.213\pm0.269$ & $32.96\pm2.60$ & $3.204\pm0.045$\\
\quad Curriculum: grid$\rightarrow$uniform & $3.405\pm0.033$ & $99.562\pm0.032$ & $93.409\pm0.314$ & $40.27\pm5.66$ & $3.268\pm0.056$\\
\quad Uniform only & $4.677\pm0.077$ & $99.180\pm0.148$ & $89.986\pm0.701$ & $108.46\pm17.00$ & $3.637\pm0.071$\\
\quad Curriculum: logit-normal$\rightarrow$uniform & $4.806\pm0.071$ & $99.006\pm0.178$ & $88.911\pm0.513$ & $133.00\pm14.32$ & $3.703\pm0.040$\\
\quad SimpleFold-style & $5.391\pm0.025$ & $99.171\pm0.078$ & $89.163\pm0.945$ & $141.95\pm25.92$ & $3.697\pm0.104$\\
\addlinespace[2pt]
\multicolumn{6}{@{}l}{\textit{Training time-grid resolution}}\\
\quad $K_{\mathrm{train}}=10$ & $3.097\pm0.023$ & $99.625\pm0.026$ & $95.195\pm0.102$ & $24.61\pm2.18$ & $3.030\pm0.021$\\
\quad $K_{\mathrm{train}}=20$ & $3.092\pm0.019$ & $99.636\pm0.011$ & $95.278\pm0.165$ & $25.07\pm2.71$ & $3.066\pm0.039$\\
\quad $K_{\mathrm{train}}=40$ & $3.089\pm0.024$ & $99.636\pm0.019$ & $95.243\pm0.086$ & $24.82\pm2.15$ & $3.091\pm0.042$\\
\quad $K_{\mathrm{train}}=100$ & $3.102\pm0.011$ & $99.614\pm0.018$ & $95.192\pm0.210$ & $23.71\pm3.14$ & $3.049\pm0.020$\\
\addlinespace[2pt]
\multicolumn{6}{@{}l}{\textit{ESM-3 conditioning}}\\
\quad Without ESM-3 residue and BOS features & $3.770\pm0.052$ & $99.564\pm0.012$ & $90.554\pm0.313$ & $70.32\pm4.27$ & $3.399\pm0.023$\\
\addlinespace[2pt]
\multicolumn{6}{@{}l}{\textit{Anchor stride}}\\
\quad Stride 1 & $2.888\pm0.042$ & $64.296\pm2.231$ & $94.556\pm0.236$ & $8.45\pm0.99$ & $3.137\pm0.075$\\
\quad Stride 2 & $3.161\pm0.032$ & $99.328\pm0.064$ & $82.955\pm0.974$ & $42.48\pm7.42$ & $3.924\pm0.042$\\
\quad Stride 8 & $3.254\pm0.020$ & $99.728\pm0.010$ & $94.133\pm0.264$ & $59.50\pm7.90$ & $3.233\pm0.034$\\
\quad Stride 16 & $3.906\pm0.008$ & $99.734\pm0.016$ & $92.621\pm0.097$ & $120.76\pm5.63$ & $3.370\pm0.023$\\
\bottomrule
\end{tabular}%
}
\end{table}

\paragraph{Time sampling.}
The selected sampler draws equal proportions from the exact source endpoint
$t=0$, the discrete Euler grid, and the continuous uniform distribution. It
has the lowest validation direct mean and median RMSD among the sampled time
distributions. Grid-only reaches a direct mean/median RMSD of 3.165/1.644~\AA,
versus 3.083/1.595~\AA{} for the selected sampler, and also has lower
pseudo-angle validity and a higher MolProbity score. Uniform-only and
SimpleFold-style sampling increase direct mean RMSD by 1.593 and 2.308~\AA{},
respectively. This result suggests that
explicit exposure to the source endpoint and inference-grid times is useful
for this conditional transport problem.

\paragraph{Training time-grid resolution.}
Here the training-grid resolution, $K_{\mathrm{train}}$, controls the support
$\{0,1/K_{\mathrm{train}},\ldots,(K_{\mathrm{train}}-1)/K_{\mathrm{train}}\}$
of the discrete component of the training-time sampler. It is not the number
of integration steps used at evaluation, which remains five for every row.
The selected $K_{\mathrm{train}}=5$ has the lowest direct mean RMSD (3.083~\AA);
$K_{\mathrm{train}}=20$ and 40 give 3.092 and 3.089~\AA, respectively,
differences that are small relative to seed variation. A grid of
10 gives a slightly lower MolProbity score (3.030 versus 3.044). We therefore
interpret this ablation as showing no consistent benefit from a finer training
grid, rather than as evidence that every metric is uniquely optimized at five.

\paragraph{ESM-3 conditioning.}
Removing both residue-level and BOS ESM-3 features increases direct mean RMSD by
0.687~\AA, decreases pseudo-angle validity by 4.71 percentage points, and
increases the MolProbity score by 0.355. The C$\alpha$ clash rate also rises
from 25.18 to 70.32 per million eligible pairs. Thus sequence-level context
supports both conformational prediction and structural fidelity. This
comparison ablates the two ESM-3 inputs together and does not isolate their
individual contributions.

\paragraph{Anchor stride.}
Stride 1 predicts every residue in the Cartesian top flow and obtains the
lowest direct mean RMSD, 2.888~\AA. It removes the recursive local-coordinate stage,
however, and its adjacent-distance validity falls from 99.63\% to 64.30\%.
Stride 8 and stride 16 preserve adjacent distances but increase global error,
clashes, and all-atom MolProbity score. Stride 2 gives lower pseudo-angle and
all-atom validity than stride 4. We therefore select stride 4 as the best
observed balance between global accuracy and local and all-atom validity, not
as the optimum for RMSD alone. Stride 1 also uses the legacy execution path
with state normalization disabled, so it is an architectural boundary case
rather than a perfectly isolated stride-only perturbation.

Overall, the validation results support the equal three-way time mixture,
ESM-3 conditioning, a five-point training grid, and stride 4 for the final
model. The uncertainty values are
seed-level sample standard deviations, not confidence intervals. We perform no
significance test and avoid interpreting small differences as statistically
resolved.

\clearpage
\section{\texorpdfstring{\rev{Local-Geometry Ablations}}{Local-Geometry Ablations}}
\label{app:mechanistic_controls}

\rev{These controls were performed after the test-reported configuration had
been fixed and use only the 150-pair validation split. They are explanatory
rather than a second round of test-set model selection. Five-seed radius/loss
controls use seeds $\{1,3,42,88,99\}$; the representation and
auxiliary/projection controls use matched seeds $\{1,3,42\}$. C$\alpha$
statistics come from the original validation predictions (150 pairs per seed)
and use 55,064 within-segment adjacent pairs and 54,571 contiguous triplets.
Fixed-C$\alpha$ cg2all and MolProbity use the identical 149 successful pairs
per seed; the excluded pair is \texttt{4P7G\_C\_3OE4\_A}. Values are means and
seed-level sample standard deviations, not confidence intervals.}

\begin{table}[htbp]
\caption{Fixed-radius and distance-loss controls on validation. Values are
mean $\pm$ sample SD over five seeds. Fixed denotes the reported apo-adjacent
and final-leaf radius rules, not fixed radii everywhere. All rows retain
coordinate and feasibility losses and non-final feasibility projection.
C$\alpha$ metrics use 150 pairs per seed and direct RMSD is measured in the shared pre-aligned D3PM frame without a post-hoc target fit; MolProbity (MP) uses the identical
149 successful pairs.}
\label{tab:radius_loss_controls}
\centering
\scriptsize
\setlength{\tabcolsep}{3.3pt}
\resizebox{\linewidth}{!}{%
\begin{tabular}{lrrrrr}
\toprule
Setting & Direct mean RMSD (\AA) $\downarrow$ & Dist. valid (\%) $\uparrow$
& Angle valid (\%) $\uparrow$ & C$\alpha$ clash/$10^6$ $\downarrow$
& MP $\downarrow$\\
\midrule
Fixed apo/final-leaf radii, full losses
& $3.083\pm0.011$ & $99.632\pm0.021$ & $95.261\pm0.225$
& $25.18\pm3.73$ & $3.044\pm0.064$\\
Learned all radii, full losses
& $3.072\pm0.016$ & $99.482\pm0.019$ & $95.319\pm0.152$
& $24.95\pm1.77$ & $3.009\pm0.015$\\
Fixed apo/final-leaf radii, no adjacent loss
& $3.074\pm0.034$ & $99.634\pm0.026$ & $95.364\pm0.178$
& $24.08\pm4.09$ & $3.064\pm0.050$\\
Learned all radii, no adjacent loss
& $3.075\pm0.027$ & $99.472\pm0.019$ & $95.396\pm0.168$
& $24.51\pm2.56$ & $3.032\pm0.035$\\
Fixed apo/final-leaf radii, no distance losses
& $3.075\pm0.019$ & $99.635\pm0.016$ & $95.311\pm0.170$
& $25.07\pm1.97$ & $3.076\pm0.033$\\
Learned all radii, no distance losses
& $3.071\pm0.025$ & $99.475\pm0.018$ & $95.435\pm0.103$
& $22.95\pm2.78$ & $2.990\pm0.037$\\
\bottomrule
\end{tabular}}
\end{table}

\begin{table}[htbp]
\caption{Three-seed local-representation and auxiliary/projection controls on
validation. RMSD is direct in the shared pre-aligned D3PM frame. Minimal Local ICS and Local Cartesian disable fixed radii, all
geometric auxiliary losses, and feasibility projection. Local ICS +
auxiliaries/projection retains coordinate and feasibility losses and the
non-final projection. All rows learn every radius, disable both distance
losses, and use the same flow-training schedule.}
\label{tab:three_seed_controls}
\centering
\scriptsize
\setlength{\tabcolsep}{3.3pt}
\resizebox{\linewidth}{!}{%
\begin{tabular}{lrrrrrr}
\toprule
Setting & Median direct RMSD (\AA) $\downarrow$ & Mean direct RMSD (\AA) $\downarrow$
& Dist. valid (\%) $\uparrow$ & Angle valid (\%) $\uparrow$
& C$\alpha$ clash/$10^6$ $\downarrow$ & MP $\downarrow$\\
\midrule
Minimal Local ICS
& $1.559\pm0.010$ & $3.064\pm0.014$ & $99.486\pm0.018$
& $95.207\pm0.049$ & $26.16\pm2.28$ & $3.025\pm0.042$\\
Local Cartesian
& $1.601\pm0.091$ & $3.037\pm0.010$ & $91.735\pm0.459$
& $96.364\pm0.259$ & $23.07\pm5.69$ & $2.931\pm0.042$\\
Local ICS + auxiliaries/projection
& $1.565\pm0.028$ & $3.060\pm0.026$ & $99.486\pm0.010$
& $95.449\pm0.141$ & $21.83\pm1.84$ & $2.980\pm0.043$\\
\bottomrule
\end{tabular}}
\end{table}

\begin{table}[htbp]
\caption{All-atom details for the three-seed controls in
Table~\ref{tab:three_seed_controls}. MolProbity evaluates the same 149
relaxation-free cg2all reconstructions per seed.}
\label{tab:three_seed_allatom}
\centering
\scriptsize
\setlength{\tabcolsep}{3.5pt}
\resizebox{\linewidth}{!}{%
\begin{tabular}{lrrrrr}
\toprule
Setting & Clashscore $\downarrow$ & Rama favored (\%) $\uparrow$
& Rama outlier (\%) $\downarrow$ & Bond RMSD (\AA) $\downarrow$
& Angle RMSD (deg) $\downarrow$\\
\midrule
Minimal Local ICS
& $95.44\pm5.73$ & $90.174\pm0.188$ & $3.319\pm0.020$
& $0.02484\pm0.00051$ & $2.866\pm0.031$\\
Local Cartesian
& $94.51\pm6.64$ & $91.355\pm0.185$ & $3.086\pm0.099$
& $0.03188\pm0.00061$ & $3.327\pm0.065$\\
Local ICS + auxiliaries/projection
& $90.18\pm4.45$ & $90.404\pm0.399$ & $2.992\pm0.162$
& $0.02423\pm0.00067$ & $2.840\pm0.082$\\
\bottomrule
\end{tabular}}
\end{table}

\rev{\paragraph{Fixed radii and distance losses.}
Table~\ref{tab:radius_loss_controls} compares the reported fixed
apo-adjacent/final-leaf radius rules with learned-all-radii variants, then
removes the adjacent-distance loss or both adjacent and sparse-distance losses.
Feasibility projection remains enabled in all six rows. Learning all radii
under the full objective lowers distance fidelity by only 0.150 percentage
points ($99.632\%$ to $99.482\%$), while direct mean RMSD and MolProbity slightly
improve. Removing either set of distance losses changes the metrics only
weakly and inconsistently relative to seed variation. The fixed-radius rule
and distance auxiliaries therefore act as minor refinements rather than the
primary source of local fidelity.}

\rev{\paragraph{Local representation.}
The representation control retains the same stride-4 hierarchy, generated
parent-anchor frame, apo reference, network capacity, conditioning, and
training schedule. Both variants learn every radius and disable all geometric
auxiliary losses and infeasibility projection. Local Cartesian predicts the
three components of $\mathbf{M}-(\mathbf{A}+\mathbf{B})/2$ in the same $(\mathbf{e}_1,\mathbf{e}_2,\mathbf{e}_3)$ parent frame,
padded by one zero channel to preserve the model interface. It has lower direct mean RMSD but higher direct median RMSD, higher
pseudo-angle validity, and lower MolProbity. Local
ICS, however, raises adjacent-distance fidelity by 7.751 percentage points and
reduces reconstructed bond and angle RMSDs. The shared frame defines each new
midpoint relative to its generated parent anchors and the apo reference; the
ICS distance parameterization adds a stronger bond-scale geometric bias. This
supports a specific role in local geometry, not uniform superiority on every
metric.}

\rev{\paragraph{Other geometric auxiliaries and projection.}
Minimal Local ICS uses the same local-ICS architecture as the reported model,
but learns every radius, removes all four geometric auxiliary losses
($\mathcal L_{\mathrm{xyz}}$, $\mathcal L_{\mathrm{adj}}$,
$\mathcal L_{\mathrm{sparse}}$, and $\mathcal L_{\mathrm{feas}}$), and disables
non-final sphere-feasibility projection. The \emph{Local ICS +
auxiliaries/projection} control also learns every radius and disables the two
distance losses, but retains coordinate and feasibility losses and the
non-final projection. Their direct mean RMSD and adjacent-distance fidelity are
essentially unchanged. Removing these terms modestly worsens pseudo-angle
validity, C$\alpha$ clashes, and MolProbity, indicating that they are mild
regularizers rather than the primary source of accuracy or local fidelity.}

\clearpage
\section{Full-Chain Internal Coordinates and Lever-Arm Diagnostics}
\label{app:lever_arm}

The shallow midpoint hierarchy does not make generated-parent errors vanish.
Instead, each local prediction depends on one parent interval, whereas a
sequential full-chain internal-coordinate rollout can transmit an early
angular error through every downstream reconstruction. We examine this
mechanism with two previously trained full-chain controls on the fixed
150-pair D3PM test split and with a controlled reconstruction diagnostic.

\begin{table}[H]
\centering
\scriptsize
\setlength{\tabcolsep}{4.5pt}
\resizebox{\linewidth}{!}{%
\begin{tabular}{lrrrr}
\toprule
Model & Median direct RMSD (\AA) $\downarrow$ & Mean direct RMSD (\AA) $\downarrow$
& Dist. $\leq0.2$~\AA{} (\%) $\uparrow$ & Angle $\leq20^\circ$ (\%) $\uparrow$\\
\midrule
Full-chain ICS Transformer & 7.559 & 10.377 & \textbf{99.699} & \textbf{98.833}\\
Full-chain ICS NeRF--EGNN & 7.618 & 10.489 & \textbf{99.699} & 98.793\\
\method{} (seed 42) & \textbf{1.338} & \textbf{2.780}
& 99.587 & 95.599\\
\bottomrule
\end{tabular}
}
\caption{Diagnostic comparison on the fixed 150-pair D3PM test split. The
full-chain controls sequentially reconstruct the C$\alpha$ trace from global
internal coordinates. Because their NeRF root pose is undefined, we align each
decoded trace to the pre-aligned apo input and then compute RMSD to holo without
a prediction-to-holo fit. Local metrics pool 54,509 within-segment adjacent
pairs and 54,082 contiguous triplets. All rows use seed 42, but the controls
are not parameter-matched ablations; they provide supporting rather than
causal evidence for the hierarchy.}
\label{tab:full_ics_controls}
\end{table}

\rev{This conclusion also replicates on the separate fixed 150-pair D3PM
validation split using the same seed-42 checkpoints and metric protocols. The
full-chain ICS Transformer, full-chain ICS NeRF--EGNN, and \method{} obtain
median/mean direct RMSDs of $8.900/11.677$, $8.300/11.396$, and
$1.632/3.074$~\AA, respectively. Their adjacent-distance/pseudo-angle
fidelities are $99.809/98.824\%$, $99.809/98.668\%$, and
$99.599/94.891\%$, respectively. Validation local metrics pool 55,064
within-segment adjacent pairs and 54,571 contiguous triplets.}

\begin{figure}[H]
\centering
\includegraphics[width=0.58\linewidth]{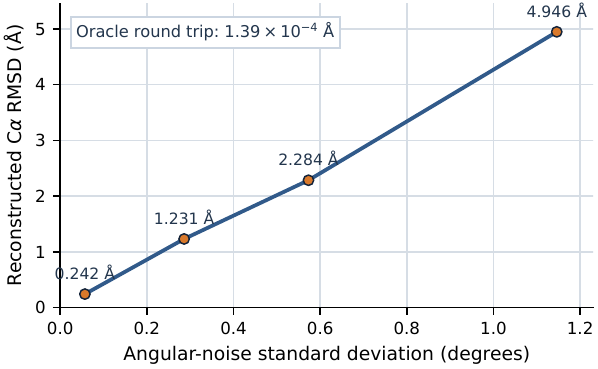}
\caption{Controlled full-chain reconstruction diagnostic. Sub-degree angular
noise produces increasing Cartesian displacement despite fixed distance
channels and a numerically exact oracle round trip.}
\label{fig:lever_arm_noise}
\end{figure}

The full-chain models match or exceed \method{} on the local threshold metrics
yet have substantially larger direct RMSD. This decoupling is consistent with
the lever-arm effect: small local angular errors can accumulate into large
downstream Cartesian displacements. To separate this effect from a decoder implementation failure,
an oracle encode--decode round trip gives $1.39\times10^{-4}$~\AA{} RMSD.
We then add zero-mean Gaussian noise to the angular channels of a 265-residue
trace while keeping its distance channels fixed. Angular standard deviations
of $0.001$, $0.005$, $0.010$, and $0.020$ radians
($0.057^\circ$, $0.286^\circ$, $0.573^\circ$, and $1.146^\circ$) produce
Cartesian RMSDs of $0.242$, $1.231$, $2.284$, and $4.946$~\AA,
respectively (Figure~\ref{fig:lever_arm_noise}). Upstream torsions have up to
$328.8\times$ larger coordinate sensitivity than the least influential
torsions. These diagnostics show chain-length amplification in sequential ICS;
they do not claim that the hierarchical model eliminates parent-error
propagation.

\clearpage
\section{Training and Inference Algorithm}
\label{app:generation_algorithm}

Algorithm~\ref{alg:crypticflow_inference} makes explicit the different contexts
used during training and generation. During training, the top-anchor module
regresses the clean holo endpoint from interpolated apo--holo anchors. The
midpoint module is trained level by level with ground-truth parent coordinates
and clean local-coordinate targets. This teacher-forced construction prevents
errors from earlier generated levels from changing the target distribution
seen by the midpoint Transformer during optimization. The adjacent-distance
rule is also active during training: radius channels next to a
sequence-adjacent parent are fixed to the sanitized apo distance before
decoding. Non-final decoded states use the same sphere-feasibility repair as inference,
and their Cartesian coordinates contribute to geometric auxiliary losses, including
an adjacent-distance term.

At inference, the two stages are composed autoregressively across hierarchy
levels. The top-anchor flow is integrated first, and its generated anchors
initialize a partial C$\alpha$ trace. At each midpoint level, all active
intervals are processed in parallel from fresh Gaussian local states. The
integrated states are decoded with the adjacent-distance and sphere-feasibility
rules, merged into the partial trace, and reused as parents for the next level.
The procedure is therefore sequential across levels but parallel within each
level. For stride four, two midpoint levels complete every full interval.
\begin{algorithm}[H]
\caption{Training and inference with hierarchical flow matching.}
\label{alg:crypticflow_inference}
\begin{minipage}{0.97\linewidth}
\scriptsize
\renewcommand{\arraystretch}{0.95}
\textbf{Model components:}\par\vspace{-1pt}
\begin{tabular}{@{}p{0.115\linewidth}@{\hspace{0.6em}}p{0.82\linewidth}@{}}
$f_{\theta}^{\mathrm{top}}$ & ESM-3-conditioned E(3)-equivariant anchor EGNN;\\
$f_{\theta}^{\mathrm{mid}}$ & conditional midpoint Transformer;\\
$\mathcal D_{\mathrm{ICS}}$ & local-ICS geometry decoder and merge operator;\\
$\operatorname{Euler}_{K}$ & clean-endpoint flow integration with $K=5$ steps.
\end{tabular}

\vspace{2pt}\textbf{Stage I: Data preparation}\par\vspace{-1pt}
\hspace{0.8em}\begin{tabular}{@{}r@{\hspace{0.7em}}p{0.84\linewidth}@{}}
1: & Split each paired trace into contiguous segments; select stride-4 anchors $\mathcal A$ and recursive midpoint levels $\{\mathcal M_\ell\}_{\ell=1}^{D}$.\\
2: & Build residue features $\mathbf{F}$, apo-referenced local frames, and clean local targets $\mathbf{S}_1$.
\end{tabular}

\vspace{2pt}\textbf{Stage II: Training}\par\vspace{-1pt}
\hspace{0.8em}\begin{tabular}{@{}r@{\hspace{0.7em}}p{0.84\linewidth}@{}}
3: & Sample $(\mathbf{X}_{\mathrm{apo}},\mathbf{X}_{\mathrm{holo}},\mathbf{F})$ and flow time $t$.\\
4: & Set $\mathbf{X}_t=(1-t)\mathbf{X}_{\mathrm{apo},\mathcal A}+t\mathbf{X}_{\mathrm{holo},\mathcal A}$ and regress $f_{\theta}^{\mathrm{top}}(\mathbf{X}_t,t,\mathbf{F})$ to $\mathbf{X}_{\mathrm{holo},\mathcal A}$.\\
5: & \textbf{for} level $\ell=1,\ldots,D$ with teacher-forced parent coordinates \textbf{do}\\
6: & \quad Draw $\mathbf{S}_0\sim\mathcal N(0,\mathbf{I})$ and set $\mathbf{S}_t=(1-t)\mathbf{S}_0+t\mathbf{S}_1$.\\
7: & \quad Regress $f_{\theta}^{\mathrm{mid}}(\mathbf{S}_t,t,\ell,\mathbf{F},\mathrm{parents})$ to $\mathbf{S}_1$.\\
8: & \quad Apply adjacent-distance and sphere-feasibility rules to the decoded training state; evaluate geometric auxiliary losses.\\
9: & \textbf{end for}\\
10: & Update $\theta$ with the joint clean-endpoint and geometric objective.
\end{tabular}

\vspace{2pt}\textbf{Stage III: Inference}\par\vspace{-1pt}
\hspace{0.8em}\begin{tabular}{@{}r@{\hspace{0.7em}}p{0.84\linewidth}@{}}
11: & Generate top anchors $\mathbf{X}_{\mathcal A}\leftarrow\operatorname{Euler}_{K}(f_{\theta}^{\mathrm{top}},\mathbf{X}_{\mathrm{apo},\mathcal A},\mathbf{F})$.\\
12: & Initialize the partial trace $\mathcal P\leftarrow \mathbf{X}_{\mathcal A}$.\\
13: & \textbf{for} level $\ell=1,\ldots,D$ \textbf{do}\\
14: & \quad Build unresolved parent intervals and draw fresh $\mathbf{S}_0\sim\mathcal N(0,\mathbf{I})$.\\
15: & \quad Set $\widehat{\mathbf S}_1\leftarrow\operatorname{Euler}_{K}(f_{\theta}^{\mathrm{mid}},\mathbf{S}_0,\ell,\mathbf{F},\mathcal P)$.\\
16: & \quad Apply adjacent-distance and sphere-feasibility rules; update $\mathcal P\leftarrow\mathcal P\cup\mathcal D_{\mathrm{ICS}}(\widehat{\mathbf S}_1)$.\\
17: & \textbf{end for}\\
18: & \textbf{return} the completed holo-like C$\alpha$ trace $\widehat{\mathbf X}_1\leftarrow\mathcal P$.
\end{tabular}
\end{minipage}
\end{algorithm}
\newsavebox{\revtablebox}
\newcommand{\revfit}[1]{%
  \sbox\revtablebox{#1}%
  \ifdim\wd\revtablebox>\linewidth
    \resizebox{\linewidth}{!}{\usebox\revtablebox}%
  \else
    \usebox\revtablebox
  \fi}

\section{\rev{External Generalization: CryptoBench}}
\label{app:cryptobench}

\rev{The results in the main text are confined to the fixed D3PM-Large split.
To test generalization beyond it, we applied the models trained on D3PM-Large,
without any retraining, to CryptoBench \citep{skrhak2024cryptobench}.}

\paragraph{\rev{Leakage filter and evaluation set.}}
\rev{To remove proteins overlapping the training data we compared \emph{both}
the apo and the holo structure of every CryptoBench record against the apo and
holo structures of all 1,591 D3PM-Large pairs (train, validation and test). A
record was excluded if any of the following held: (i) the same PDB identifier
and chain, or the same UniProt accession; (ii) full-length sequence identity
$\geq30\%$ with $\geq80\%$ coverage of both sequences (MMseqs2); (iii)
domain-level similarity with identity $\geq30\%$, alignment length $\geq80$
residues, $\geq80\%$ coverage of the shorter sequence and E-value
$\leq10^{-3}$. Of the 5,493 original records, 2,517 (650 distinct apo PDB
entries) pass this filter together with the structural checks, which require at
least 50 matched C$\alpha$ atoms between apo and holo, at least 5 pocket
C$\alpha$ atoms, and a holo ligand. D3PM-Large was curated to contain only
pairs whose apo--holo C$\alpha$ RMSD exceeds 3~\AA{}
\citep{minan2025sesame}. Applying the same curation criterion to CryptoBench
leaves \textbf{212 records over 59 distinct apo PDB entries}. The selection
uses only apo and holo structures; no prediction was consulted. The record list
is released with the code.}

\paragraph{\rev{Metrics and weighting.}}
\rev{Global C$\alpha$ RMSD is the direct RMSD in the shared pre-aligned frame,
as in the main text. Distance and pseudo-angle fidelity are as in the main
text. MolProbity is computed after the same fixed-C$\alpha$ cg2all
reconstruction used for every method, on the 211 records (58 apo PDB entries)
for which reconstruction and scoring succeeded for every structure. Because one
apo structure can carry several ligand records, every metric is first averaged
within an apo PDB entry and then averaged with equal weight over apo PDB
entries.}

\begin{table}[h]
\caption{\rev{CryptoBench, apo--holo C$\alpha$ RMSD $\geq3$~\AA{} (212 records,
59 apo PDB entries). All methods were trained on the D3PM-Large training split
with five seeds and applied without retraining. Mean $\pm$ sample standard
deviation over seeds.}}
\label{tab:cryptobench}
\centering
\small
\revfit{%
\begin{tabular}{lcccc}
\toprule
Method & Global C$\alpha$ RMSD (\AA) $\downarrow$ & Distance fidelity (\%) $\uparrow$ & Angle fidelity (\%) $\uparrow$ & MolProbity $\downarrow$\\
\midrule
Apo (input, unmoved) & 4.23 & 99.88 & 97.29 & 2.26\\
\method{} & \textbf{4.06 $\pm$ 0.02} & \textbf{99.67 $\pm$ 0.04} & \textbf{92.28 $\pm$ 0.14} & \textbf{3.46 $\pm$ 0.04}\\
FDBM & 5.29 $\pm$ 0.14 & 25.64 $\pm$ 1.42 & 77.33 $\pm$ 1.83 & 4.34 $\pm$ 0.06\\
ABM & 5.32 $\pm$ 0.11 & 25.75 $\pm$ 0.39 & 77.06 $\pm$ 0.61 & 4.37 $\pm$ 0.02\\
SBAlign & 5.06 $\pm$ 0.04 & 11.28 $\pm$ 0.34 & 45.71 $\pm$ 0.15 & 4.65 $\pm$ 0.01\\
EGNN & 15.64 $\pm$ 0.19 & 6.25 $\pm$ 0.17 & 31.12 $\pm$ 0.73 & 4.94 $\pm$ 0.01\\
\bottomrule
\end{tabular}}
\end{table}

\rev{On this external dataset \method{} is the only evaluated method that
improves on the apo input in global C$\alpha$ RMSD; all four compared
generative models move further from holo than the unmoved apo structure does.
Among the generative models \method{} also attains the highest distance and
angle fidelity and the best MolProbity score. The apo input nevertheless leads
on local geometry and MolProbity: a structure that was not moved cannot break
its own geometry. The claim supported here is a relative advantage among models
that move the structure, together with an improvement over the apo baseline in
global accuracy.}

\subsection{\rev{Pocket Recovery and Downstream Docking}}
\label{app:cryptobench_pocket}

\rev{The main text notes that D3PM does not evaluate pocket recovery or docking
utility. Within the regime above we measured both. The pocket is the set of
C$\alpha$ atoms within 8~\AA{} of a holo ligand heavy atom, and values use the
same within-apo-then-across-apo weighting.}

\begin{table}[h]
\caption{\rev{Pocket recovery on CryptoBench (212 records, 59 apo PDB entries).
Direct RMSD, mean $\pm$ seed standard deviation over five seeds. MolProbity is
measured on the 211 records (58 apo PDB entries) on which apo, GT holo and all
five models succeeded.}}
\label{tab:cryptobench_pocket}
\centering
\small
\revfit{%
\begin{tabular}{lccc}
\toprule
Method & Global C$\alpha$ RMSD (\AA) $\downarrow$ & Pocket C$\alpha$ RMSD (\AA) $\downarrow$ & MolProbity $\downarrow$\\
\midrule
Apo (input, baseline) & 4.227 & 4.124 & 2.256\\
GT holo (reference) & --- & --- & 2.209\\
\method{} & \textbf{4.059 $\pm$ 0.021} & \textbf{3.816 $\pm$ 0.023} & \textbf{3.461 $\pm$ 0.035}\\
FDBM & 5.286 $\pm$ 0.141 & 4.928 $\pm$ 0.192 & 4.341 $\pm$ 0.060\\
ABM & 5.318 $\pm$ 0.108 & 5.027 $\pm$ 0.230 & 4.367 $\pm$ 0.021\\
SBAlign & 5.065 $\pm$ 0.043 & 4.860 $\pm$ 0.047 & 4.653 $\pm$ 0.007\\
EGNN & 15.638 $\pm$ 0.189 & 14.226 $\pm$ 0.257 & 4.943 $\pm$ 0.008\\
\bottomrule
\end{tabular}}
\end{table}

\rev{\method{} is the only method better than the apo input in both global and
pocket RMSD, and the improvement is larger at the pocket ($-0.307$~\AA) than
globally ($-0.168$~\AA): the model does not displace the protein uniformly but
moves the ligand-binding site more.}

\paragraph{\rev{Docking.}}
\rev{Within the same regime the docking targets are the 154 records over 37 apo
PDB entries whose ligand has at least 12 and at most 45 heavy atoms and is
fully resolved. Docking conditions are identical for every receptor: known-site
docking with \texttt{gnina v1.3.3}, one box fixed per record and shared
unchanged by every receptor (the bounding box of the holo ligand heavy atoms
plus 4~\AA{} on each face). Every model receptor is a predicted C$\alpha$ trace
converted to all atoms by the same cg2all reconstruction, and the ligand is a
conformer freshly embedded with RDKit ETKDGv3; the crystal pose is never used
as input. Pose RMSD is symmetry-corrected and measured without refitting.}

\begin{table}[h]
\caption{\rev{Known-site docking on CryptoBench (154 records, 37 apo PDB
entries). top-1 is the pose ranked first by CNN rescoring; best-of-20 is the
closest of the 20 generated poses.}}
\label{tab:cryptobench_docking}
\centering
\small
\revfit{%
\begin{tabular}{lccc}
\toprule
Receptor & Median top-1 pose RMSD (\AA) $\downarrow$ & Median best-of-20 (\AA) $\downarrow$ & 2~\AA{} success $\uparrow$\\
\midrule
Holo crystal structure (positive control) & 2.147 & 1.175 & 50.2\%\\
GT holo backbone, reconstructed (ceiling) & 4.170 & 2.908 & 11.1\%\\
Apo (input) & 6.473 & 4.642 & 2.7\%\\
\method{} & \textbf{5.781} & \textbf{4.360} & \textbf{5.4\%}\\
ABM & 6.493 & 4.730 & 0.0\%\\
SBAlign & 6.862 & 4.844 & 0.0\%\\
FDBM & 7.026 & 4.837 & 0.0\%\\
EGNN & 8.344 & 5.979 & 0.0\%\\
\bottomrule
\end{tabular}}
\end{table}

\rev{With a \method{} prediction as the receptor, both top-1 and best-of-20 are
closer to the correct pose than with the apo input or with any of the other
four generative models, and \method{} is the only generative model that
produced any pose meeting the 2~\AA{} criterion. Two controls set the scope.
The ceiling, which passed through the same cg2all reconstruction, reaches only
an 11.1\% success rate at 2~\AA. Getting the backbone right is therefore not
sufficient for C$\alpha$-level prediction to recover most of the docking
success; side-chain reconstruction is the remaining bottleneck, and the gap to
the crystal structure (50.2\%) measures that shortfall. This result shows that
improvements in backbone accuracy propagate into smaller ligand-placement
errors; it does not claim to solve docking success itself.}

\section{\rev{Capacity- and Pretraining-Matched Baselines}}
\label{app:capacity}

\rev{Section~\ref{sec:discussion} notes that \method{} is substantially larger
than the published bridge baselines, so the main comparison is not capacity
matched. We therefore retrained all four baselines from scratch at \method{}'s
parameter count and gave them the same ESM-3 features as input. Data (the
D3PM-Large training split), epoch budget (1,000), effective batch size (8) and
checkpoint-selection criterion (validation direct C$\alpha$ RMSD) were matched.
Parameter counts measured from the checkpoint state dictionaries are
\method{} 34{,}080{,}122, ABM and FDBM 33{,}693{,}525, SBAlign 33{,}751{,}329
and EGNN 35{,}072{,}991, that is within $\pm3\%$ of each other.}

\begin{table}[h]
\caption{\rev{Capacity- and ESM-3-matched baselines on the fixed 150-pair D3PM
test split. Direct RMSD. \method{} is the mean $\pm$ sample standard deviation
over five seeds; every baseline is a single seed. Bold marks the best among the
models that generate structures; apo and holo are reference lines.}}
\label{tab:capacity_internal}
\centering
\small
\revfit{%
\begin{tabular}{lccccccc}
\toprule
Method & Params & RMSD (\AA) $\downarrow$ & Median (\AA) $\downarrow$ & $<2$~\AA{} (\%) $\uparrow$ & Dist.\ fid.\ (\%) $\uparrow$ & Angle fid.\ (\%) $\uparrow$ & MolProbity $\downarrow$\\
\midrule
holo (upper bound) & --- & 0 & 0 & 100 & 100 & 100 & 2.291\\
apo (unchanged) & --- & 6.743 & 4.461 & 0.00 & 99.70 & 95.59 & 2.392\\
\method{} & 34.08M & 2.766 $\pm$ 0.041 & 1.333 $\pm$ 0.056 & 58.53 $\pm$ 0.73 & \textbf{99.58} & \textbf{95.87} & \textbf{3.034 $\pm$ 0.054}\\
ABM & 33.69M & \textbf{2.547} & \textbf{1.069} & \textbf{68.00} & 38.46 & 90.21 & 3.802\\
FDBM & 33.69M & 4.190 & 3.328 & 7.33 & 37.73 & 84.74 & 3.962\\
SBAlign & 33.75M & 3.675 & 2.467 & 22.00 & 11.18 & 47.10 & 4.612\\
EGNN & 35.07M & 4.189 & 2.614 & 38.00 & 16.82 & 59.88 & 4.388\\
\bottomrule
\end{tabular}}
\end{table}

\paragraph{\rev{The accuracy claim does not survive capacity matching.}}
\rev{Once the parameter count is matched, ABM attains a lower global RMSD than
\method{} (2.547 versus 2.766~\AA), and also leads on the median (1.069 versus
1.333~\AA) and on the sub-2~\AA{} success rate (68.00 versus 58.53\%). Our
seed-to-seed standard deviation is 0.041~\AA, so this is not seed noise. The
claim in Table~\ref{tab:accuracy} that \method{} attains the lowest compared
median and mean RMSD therefore holds at the published baseline capacities and
not in general. What the same ABM does not reach is structural quality: its
adjacent-distance fidelity of 38.46\% is less than half of our 99.58\%, its
angle fidelity is 90.21 versus 95.87\%, and its MolProbity is 3.802 versus
$3.034\pm0.054$. Inside this split every model, including all four baselines,
is better than the unmoved apo input (6.743~\AA).}

\begin{table}[h]
\caption{\rev{The same four checkpoints carried, without retraining, to
CryptoBench (212 records, 59 apo PDB entries; weighting as in
Appendix~\ref{app:cryptobench}).}}
\label{tab:capacity_external}
\centering
\small
\revfit{%
\begin{tabular}{lccccccc}
\toprule
Method & Params & RMSD (\AA) $\downarrow$ & Median (\AA) $\downarrow$ & $<2$~\AA{} (\%) $\uparrow$ & Dist.\ fid.\ (\%) $\uparrow$ & Angle fid.\ (\%) $\uparrow$ & MolProbity $\downarrow$\\
\midrule
holo (upper bound) & --- & 0 & 0 & 100 & 100 & 100 & 2.209\\
apo (unchanged) & --- & 4.227 & 3.974 & 0.00 & 99.88 & 97.29 & 2.256\\
\method{} & 34.08M & \textbf{4.059 $\pm$ 0.021} & \textbf{4.006 $\pm$ 0.053} & 1.51 $\pm$ 0.62 & \textbf{99.67 $\pm$ 0.04} & \textbf{92.28 $\pm$ 0.14} & \textbf{3.461 $\pm$ 0.035}\\
ABM & 33.69M & 4.327 & 4.742 & 1.89 & 36.53 & 88.99 & 4.037\\
FDBM & 33.69M & 4.515 & 4.515 & 0.47 & 40.56 & 89.93 & 3.890\\
SBAlign & 33.75M & 4.764 & 4.644 & 0.00 & 11.06 & 46.81 & 4.644\\
EGNN & 35.07M & 8.744 & 11.338 & 0.00 & 14.61 & 54.09 & 4.668\\
\bottomrule
\end{tabular}}
\end{table}

\paragraph{\rev{The ordering reverses outside the training distribution.}}
\rev{With the same checkpoints, ABM's global RMSD of 2.547~\AA{} on the D3PM
test becomes 4.327~\AA{} on CryptoBench while ours goes from 2.766 to
4.059~\AA. Among the generative models \method{} now leads on every column
except the sub-2~\AA{} rate, where ABM's 1.89\% lies inside our seed standard
deviation ($1.51\pm0.62\%$) and no ordering can be claimed. More tellingly,
none of the four capacity-matched baselines beats the unmoved apo input
(4.227~\AA): all four lie between 4.327 and 8.744~\AA, so moving the structure
took them further from the answer. Inside the training distribution all five
models beat the apo input; outside it that property remains only for
\method{}. The gap in local geometry is not filled by capacity either --- the
best adjacent-distance fidelity among the baselines is FDBM's 40.56\%, against
our 99.67\%, and SBAlign shows most clearly that capacity is not the
bottleneck, moving from 11.3 to 11.1\% in distance fidelity and from 4.65 to
4.64 in MolProbity as it grows from 0.54M to 34M.}

\begin{table}[h]
\caption{\rev{Training and inference cost at matched capacity, extending the
hardware-controlled comparison of Table~\ref{tab:compute}. Inference was timed
on one a5000, running the 150 test pairs sequentially. GPU-hours are normalized
to 1,000 epochs.}}
\label{tab:capacity_compute}
\centering
\small
\revfit{%
\begin{tabular}{lcccccc}
\toprule
Method & Training card & Count & GPU-h / 1,000 ep.\ $\downarrow$ & vs \method{} & 150 pairs (s) $\downarrow$ & Per pair (s) $\downarrow$\\
\midrule
\method{} & RTX 6000 Ada 48\,GB & \textbf{1} & \textbf{13.3} & \textbf{1.0$\times$} & \textbf{37.4} & \textbf{0.249}\\
EGNN & RTX 6000 Ada 48\,GB & 1 & 38.1 & 2.9$\times$ & 24.9 & 0.166\\
SBAlign & H100 80\,GB & 8 & 208 & 15.7$\times$ & 2{,}500.4 & 16.669\\
ABM & RTX 6000 Blackwell & 4 & 255 & 19.2$\times$ & 2{,}364.8 & 15.765\\
FDBM & RTX 6000 Blackwell & 4 & 255 & 19.3$\times$ & 2{,}388.9 & 15.926\\
\bottomrule
\end{tabular}}
\end{table}

\rev{The one pair comparable on identical hardware is \method{} and EGNN: both
ran 1,000 epochs on a single RTX 6000 Ada, and \method{} took 2.9$\times$ less
time. The other three do not fit on a single 48\,GB card at this size and
required multiple GPUs. Effective batch size was matched at 8 for every method.
GPU-hours mix cards of different speeds, so card names are given; since H100
and Blackwell are faster cards than the RTX 6000 Ada, the ratios are
conservative. Inference is 63$\times$ faster than ABM because the bridge family
takes 100 integration steps where \method{} takes five Euler steps per active
flow.}

\paragraph{\rev{Scope.}}
\rev{Only \method{} is run with five seeds; every capacity-matched baseline is
a single seed. MolProbity on CryptoBench is aggregated over the 211 commonly
successful pairs, excluding the one pair that fails identically for every model
(CB002742, duplicate atom labels). The fixed-C$\alpha$ reconstruction leaves
C$\alpha$ positions unchanged, so global RMSD and the two fidelity measures are
unaffected by it.}

\section{\rev{Test-Split Controls Added in Revision}}
\label{app:test_controls}

\rev{The controls in Appendix~\ref{app:mechanistic_controls} were run on the
validation split and are explanatory. The controls below were requested during
review and are run on the fixed 150-pair \emph{test} split with the same five
seeds as \method{} (1, 3, 42, 88, 99). RMSD is direct C$\alpha$ RMSD.
MolProbity and the covalent-geometry RMSDs use fixed-C$\alpha$ cg2all
reconstruction without relaxation.}

\subsection{\rev{Language-Model Features}}
\label{app:test_controls_lm}

\rev{ESM-3 was pretrained on structural as well as sequence data, so structural
information about the evaluated proteins could in principle enter through its
embeddings. We replaced the ESM-3 embeddings with those of ESM-2
(\texttt{esm2\_t33\_650M\_UR50D}), which is trained on sequences only; the
1280-dimensional embeddings were zero-padded to the 1536-dimensional input and
every other setting was kept.}

\begin{table}[h]
\caption{\rev{Language-model and representation controls on the test split.}}
\label{tab:test_controls_lm}
\centering
\small
\revfit{%
\begin{tabular}{lccccccc}
\toprule
Model & Median (\AA) $\downarrow$ & Mean (\AA) $\downarrow$ & $<2$~\AA{} (\%) $\uparrow$ & Dist.\ fid.\ (\%) $\uparrow$ & Angle fid.\ (\%) $\uparrow$ & MolProbity $\downarrow$ & Bond-len.\ RMSD (\AA) $\downarrow$\\
\midrule
\method{} & 1.333 $\pm$ 0.056 & 2.766 $\pm$ 0.041 & 58.53 $\pm$ 0.73 & 99.58 $\pm$ 0.02 & 95.87 $\pm$ 0.25 & 3.034 $\pm$ 0.054 & 0.0242 $\pm$ 0.0006\\
Local Cartesian (matched) & 1.247 $\pm$ 0.058 & 2.703 $\pm$ 0.033 & 58.93 $\pm$ 1.46 & 93.41 $\pm$ 0.51 & 97.01 $\pm$ 0.04 & 2.928 $\pm$ 0.040 & 0.0317 $\pm$ 0.0012\\
ESM-3 $\to$ ESM-2 & 1.321 $\pm$ 0.042 & 2.765 $\pm$ 0.019 & 58.00 $\pm$ 0.82 & 99.59 $\pm$ 0.01 & 96.66 $\pm$ 0.10 & 2.966 $\pm$ 0.022 & 0.0234 $\pm$ 0.0004\\
No language model & 2.507 $\pm$ 0.092 & 3.555 $\pm$ 0.048 & 43.47 $\pm$ 1.66 & 99.51 $\pm$ 0.03 & 91.67 $\pm$ 0.25 & 3.376 $\pm$ 0.041 & 0.0297 $\pm$ 0.0007\\
\bottomrule
\end{tabular}}
\end{table}

\rev{Switching to ESM-2 does not change performance (seed-paired differences:
median RMSD $-0.01\pm0.08$~\AA, sub-2~\AA{} $-0.5\pm1.1$~pp, MolProbity
$-0.07\pm0.06$, angle fidelity $+0.8\pm0.2$~pp; covalent bond-length and
bond-angle RMSDs slightly lower in all five seeds). Removing the language-model
features worsens every accuracy and quality metric in all five seeds (median
RMSD $+1.17$~\AA, sub-2~\AA{} $-15.1$~pp, angle fidelity $-4.2$~pp, MolProbity
$+0.34$). \method{} therefore relies on pretrained language-model features, but
the benefit is explained by information learned from sequences rather than by
the structural pretraining of ESM-3.}

\subsection{\rev{The Adjacent-Radius Constraint}}
\label{app:test_controls_radii}

\rev{To measure what the adjacent-radius copying contributes we trained
\emph{Predicted radii}, in which \emph{every} path by which an apo distance
enters a radius is removed: no adjacent radius is fixed, and the 1.1$\times$
apo contour-length ceiling is removed as well, so the model predicts all radii
without an upper bound. The apo structure then enters only as an input
condition, exactly as for every other model.}

\begin{table}[h]
\caption{\rev{A ladder of generation schemes on the test split, ordered by
increasing distance fidelity.}}
\label{tab:test_controls_ladder}
\centering
\small
\revfit{%
\begin{tabular}{llcccc}
\toprule
Model & Adjacent radius & Midpoint generation & Mean RMSD (\AA) $\downarrow$ & Dist.\ fid.\ (\%) $\uparrow$ & MolProbity $\downarrow$\\
\midrule
Stride 1 (pure EGNN flow) & n/a & no hierarchy, global Cartesian & 2.450 $\pm$ 0.014 & 67.34 $\pm$ 2.35 & 3.117 $\pm$ 0.087\\
Local Cartesian (matched) & n/a & hierarchical, local Cartesian & 2.703 $\pm$ 0.033 & 93.41 $\pm$ 0.51 & 2.928 $\pm$ 0.040\\
Predicted radii & predicted & hierarchical, local ICS & --- & 99.56 $\pm$ 0.01 & 3.077 $\pm$ 0.041\\
\method{} & copied from apo & hierarchical, local ICS & 2.766 $\pm$ 0.041 & 99.58 $\pm$ 0.02 & 3.034 $\pm$ 0.054\\
\bottomrule
\end{tabular}}
\end{table}

\rev{Predicting the radii instead of copying them lowers distance fidelity by
$0.02\pm0.02$~pp ($99.58\to99.56\%$) and moves MolProbity by $+0.042\pm0.051$
(seed-paired). The $99.58\%$ figure is therefore not produced by copying the
radii. The ladder shows where it does come from: a single EGNN generating all
residues directly in global Cartesian coordinates reaches 67.34\%; adding the
hierarchy with the midpoint in local Cartesian coordinates reaches 93.41\%; and
writing the midpoint in local internal coordinates reaches 99.56\% even without
any copying. The figure arises when the hierarchy and local internal
coordinates are present together, and neither alone produces it. Along the same
ladder the mean direct RMSD spans only 2.450 to 2.766~\AA{} while distance
fidelity spans 32~pp.}

\subsection{\rev{Two Regimes for the Midpoint Representation}}
\label{app:test_controls_regimes}

\rev{The representation pair of Appendix~\ref{app:mechanistic_controls} was
measured on three seeds of the validation split. We extended it to five seeds
on the test split and then carried the same checkpoints to CryptoBench.
\emph{Minimal Local ICS} is our representation alone, with the fixed
apo-derived radii, the auxiliary losses and the sphere-feasibility correction
switched off; \emph{Local Cartesian} writes the same midpoint as $(x,y,z)$ in
the same parent frame.}

\begin{table}[h]
\caption{\rev{The minimal representation pair inside and outside the training
distribution. Left: D3PM test (150 pairs). Right: CryptoBench (212 records),
with the weighting of Appendix~\ref{app:cryptobench}. Five seeds throughout.}}
\label{tab:test_controls_regimes}
\centering
\small
\revfit{%
\begin{tabular}{lccc@{\hskip 1.2em}ccc}
\toprule
& \multicolumn{3}{c}{D3PM test (in-distribution)} & \multicolumn{3}{c}{CryptoBench (out-of-distribution)}\\
\cmidrule(lr){2-4}\cmidrule(lr){5-7}
Metric & \method{} & Min.\ Local ICS & Local Cart. & \method{} & Min.\ Local ICS & Local Cart.\\
\midrule
Mean direct RMSD (\AA) $\downarrow$ & 2.766 $\pm$ 0.041 & 2.734 $\pm$ 0.021 & \textbf{2.703 $\pm$ 0.032} & 4.06 $\pm$ 0.02 & \textbf{4.05 $\pm$ 0.02} & 4.30 $\pm$ 0.13\\
Dist.\ fidelity (\%) $\uparrow$ & \textbf{99.58 $\pm$ 0.02} & 99.46 $\pm$ 0.02 & 93.17 $\pm$ 0.51 & \textbf{99.67 $\pm$ 0.04} & 99.13 $\pm$ 0.05 & 20.15 $\pm$ 1.44\\
Angle fidelity (\%) $\uparrow$ & 95.87 $\pm$ 0.25 & 96.01 $\pm$ 0.07 & \textbf{96.99 $\pm$ 0.03} & 92.28 $\pm$ 0.14 & \textbf{92.54 $\pm$ 0.17} & 57.64 $\pm$ 11.51\\
MolProbity $\downarrow$ & 3.034 $\pm$ 0.054 & 3.017 $\pm$ 0.072 & \textbf{2.924 $\pm$ 0.049} & 3.46 $\pm$ 0.04 & \textbf{3.44 $\pm$ 0.08} & 4.49 $\pm$ 0.10\\
\bottomrule
\end{tabular}}
\end{table}

\rev{Inside the training distribution the Cartesian side leads on three of the
four metrics, and adding seeds does not change that direction. We therefore do
not claim that the internal-coordinate representation is generally better in
global accuracy or all-atom quality. Outside the training distribution the
trade-off disappears: the three metrics on which Local Cartesian led all
reverse, its global RMSD of 4.30~\AA{} is worse than the unmoved apo input
(4.23~\AA), angle fidelity splits 92.54 versus 57.64\% and MolProbity 3.44
versus 4.49. The distance-fidelity gap of 6.29~pp inside becomes 79~pp outside.
The trade-off is therefore a property of the training distribution, not of the
representation. Two facts limit this reading. On external data the best
local-geometry values still belong to the apo input (distance 99.88\%, angle
97.29\%, MolProbity 2.26); the claim is that among models that move the
structure we damage it least. And Minimal Local ICS --- with the radii,
auxiliary losses and sphere correction all off --- performs essentially the
same as full \method{} here, so what generalizes outside the training
distribution is the midpoint representation rather than the rest of the
design.}

\section{\rev{Fairness of the Hierarchy-Free Controls}}
\label{app:hierarchy_fairness}

\rev{Appendix~\ref{app:lever_arm} compares \method{} against hierarchy-free
full-chain internal-coordinate controls. Review raised two objections to those
controls: that the stride-1 control used a different normalization setting, and
that the full-chain controls were smaller than \method{}. Both are correct. The
first is not a setting we chose; the second we corrected by retraining at
matched capacity, which did not change the conclusion but did narrow one of our
claims.}

\paragraph{\rev{The normalization difference is structurally forced.}}
\rev{Our state normalizer is fitted on per-channel statistics of the midpoint
internal-coordinate states, and at stride 1 the tree has depth 0, so no
midpoint is generated at all. Running the main configuration twice with only
\texttt{data.stride} changed from 4 to 1 confirms this: with normalization on
there was no state to fit and training stopped before it could begin; with it
off the run proceeded normally and logged 0 for both the generated-midpoint
count and the midpoint loss. Coordinate normalization at the top-level
Cartesian stage is on exactly as in the main configuration.}

\paragraph{\rev{Capacity-matched full-chain controls.}}
\rev{We retrained both full-chain controls at matched capacity, leaving depth
(8 layers), seed (42), data and training budget unchanged and increasing only
the width (Transformer hidden $320\to416$; NeRF--EGNN hidden $320\to384$, edge
$160\to192$).}

\begin{table}[h]
\caption{\rev{Hierarchy-free controls at matched capacity on the fixed 150-pair
D3PM test split, extending Table~\ref{tab:full_ics_controls}. The full-chain
controls copy the apo adjacent distances and predict only the angles, so their
distance fidelity is marked (= apo) and left out of that column's comparison.
Bold marks the best value in each column.}}
\label{tab:hierarchy_fairness}
\centering
\small
\revfit{%
\begin{tabular}{lcccccc}
\toprule
Setting & Params & vs \method{} & Median RMSD (\AA) $\downarrow$ & Dist.\ fid.\ (\%) & Angle fid.\ (\%) $\uparrow$ & MolProbity $\downarrow$\\
\midrule
Full-chain ICS Transformer (original) & 19.84M & 0.58$\times$ & 7.559 & 99.699 (= apo) & \textbf{98.83} & 3.142\\
\quad capacity-matched & 33.09M & 0.97$\times$ & 6.845 & 99.699 (= apo) & \textbf{98.83} & 3.174\\
Full-chain ICS NeRF--EGNN (original) & 23.67M & 0.69$\times$ & 7.618 & 99.699 (= apo) & 98.79 & 3.259\\
\quad capacity-matched & 33.79M & 0.99$\times$ & 7.151 & 99.699 (= apo) & 98.78 & 3.216\\
Stride 1 (pure EGNN flow, 5 seeds) & --- & --- & \textbf{1.153 $\pm$ 0.027} & 67.34 $\pm$ 2.35 & --- & 3.117 $\pm$ 0.087\\
\method{} (5 seeds) & 34.08M & 1.00$\times$ & 1.333 $\pm$ 0.056 & \textbf{99.58 $\pm$ 0.02} & 95.87 & \textbf{3.034 $\pm$ 0.054}\\
\bottomrule
\end{tabular}}
\end{table}

\rev{Raising capacity to 97--99\% of \method{} improves the median direct RMSD
only from 7.559 to 6.845~\AA{} and from 7.618 to 7.151~\AA, about 10\% of the
gap to \method{}; the remaining gap is still 5.1$\times$ and 5.4$\times$. The
capacity mismatch is therefore not the cause of this gap. MolProbity does not
move consistently (3.142 $\to$ 3.174 for the Transformer, 3.259 $\to$ 3.216 for
NeRF--EGNN), and distance fidelity is identical to the floating-point digits
across all four settings because that channel copies the apo distances, which
independently confirms that this figure is not model performance.}

\paragraph{\rev{What this narrows in our claim.}}
\rev{Stride 1 has a lower median RMSD than \method{} in all five seeds
(difference $0.180\pm0.037$~\AA), and in local geometry the full-chain controls
are better than we are: their pseudo-angle fidelity is 98.78--98.83\% against
our 95.87\%, and their MolProbity of 3.17--3.26 is close to our 3.034. What we
can claim is therefore not that the hierarchy improves global accuracy, nor
that it preserves local geometry better than any alternative, but that it
recovers most of the global accuracy of a Cartesian flow while losing almost no
local geometry. Against stride 1 that trade is $-0.180$~\AA{} of median RMSD
for $+32.2\pm2.3$~pp of adjacent-distance fidelity.}

\section{\rev{Exposure Bias and Robustness to the Generated Anchors}}
\label{app:exposure}

\rev{In standard training the midpoint flow sees only ground-truth parent
contexts. We applied the two remedies proposed in PAR \citep{qu2026par} and
retrained with seed 1, keeping every other setting unchanged. In
\emph{scheduled sampling} (SS) the parent context is replaced with the model's
own predictions with a probability rising from 0 to 0.5 over the first 25k of
162k steps. In \emph{noisy context learning} (NCL) isotropic Gaussian noise is
added to the ground-truth parent C$\alpha$ coordinates with probability 0.5, at
$\sigma=1.05$~\AA{} per coordinate, so that the RMSD of the noise equals the
median error of the generated anchors on the validation set (1.81~\AA).}

\begin{table}[h]
\caption{\rev{Exposure-bias remedies on the fixed 150-pair D3PM test split
(seed 1).}}
\label{tab:exposure}
\centering
\small
\revfit{%
\begin{tabular}{lcccccc}
\toprule
Model & Median (\AA) $\downarrow$ & Mean (\AA) $\downarrow$ & $<2$~\AA{} (\%) $\uparrow$ & Dist.\ fid.\ (\%) $\uparrow$ & Angle fid.\ (\%) $\uparrow$ & MolProbity $\downarrow$\\
\midrule
\method{} & 1.309 & 2.737 & 58.0 & 99.57 & 95.93 & 3.038\\
\quad $+$ SS & 1.257 & 2.750 & 59.3 & 99.54 & 95.18 & 3.121\\
\quad $+$ NCL & 1.484 & 2.810 & 59.3 & 99.40 & 94.48 & 3.131\\
\quad $+$ SS and NCL & 1.327 & 2.791 & 58.7 & 99.39 & 93.73 & 3.216\\
\bottomrule
\end{tabular}}
\end{table}

\rev{Accuracy does not improve: the median RMSD changes by $-0.05$~\AA{} with
SS and $+0.02$~\AA{} with SS and NCL, both inside our five-seed standard
deviation (0.056~\AA), and worsens by 0.17~\AA{} with NCL. Local geometry and
structure quality worsen with all three: angle fidelity drops by
0.75--2.20~pp against a seed standard deviation of 0.25~pp, and MolProbity
rises from 3.038 to 3.121--3.216.}

\paragraph{\rev{Why the mismatch does not amplify.}}
\rev{If the train--inference mismatch were the problem, the error of the final
structure would exceed the error of the anchors it was built on. On the 150
test pairs over five seeds the RMSD of the final structure is almost identical
to the RMSD measured on the anchor residues alone (median difference
$+0.013\pm0.004$~\AA). The midpoint stage therefore adds no error even when
given incorrect anchors, and the final error is set by the top-level anchor
stage. Because SS and NCL change only the training of the midpoint stage, they
cannot reduce it.}

\paragraph{\rev{Robustness to perturbed anchors.}}
\rev{We injected an isotropic Gaussian displacement $\sigma$ into the generated
top-level anchors and re-evaluated the test split; base and oracle use five
seeds, the perturbed conditions use seeds 1, 3 and 42.}

\begin{table}[h]
\caption{\rev{Anchor perturbation on the fixed 150-pair D3PM test split.}}
\label{tab:anchor_perturbation}
\centering
\small
\revfit{%
\begin{tabular}{lcc}
\toprule
Anchor condition & Anchor RMSD (\AA) & Final mean direct RMSD (\AA)\\
\midrule
Ground-truth anchors (oracle) & 0.000 & 0.775 $\pm$ 0.006\\
Generated anchors (base) & 2.628 $\pm$ 0.035 & 2.766 $\pm$ 0.041\\
$\sigma=0.25$~\AA & 2.717 $\pm$ 0.030 & 2.853 $\pm$ 0.037\\
$\sigma=0.50$~\AA & 2.905 $\pm$ 0.028 & 3.026 $\pm$ 0.035\\
$\sigma=1.00$~\AA & 3.430 $\pm$ 0.026 & 3.506 $\pm$ 0.032\\
$\sigma=2.00$~\AA & 4.732 $\pm$ 0.025 & 4.697 $\pm$ 0.030\\
\bottomrule
\end{tabular}}
\end{table}

\rev{The final structure degrades only as much as the anchors do, with no
further accumulation: in the strongest condition the anchor error rises from
2.628 to 4.732~\AA{} while the final error rises from 2.766 to 4.697~\AA, that
is, by less. Supplying ground-truth anchors lowers the final error from 2.766
to 0.775~\AA, which locates 72\% of the remaining global error at the top-level
anchor stage. The component to improve is therefore the anchor stage, not the
midpoint stage. The injected displacement is isotropic and does not reproduce
the correlation and directionality of the real anchor error, so this table is
an upper bound under isotropic perturbation rather than a reproduction of it.}

\section{\rev{Sampling Diversity}}
\label{app:diversity}

\rev{Using each of the five trained models we generated 20 samples for each of
the 150 test pairs. Diversity is the mean pairwise C$\alpha$ RMSD among the 20
samples of a target; over the 150 targets its median is $0.020\pm0.003$~\AA{}
and its 90th percentile $0.22\pm0.05$~\AA. Against a median apo--holo RMSD of
4.46~\AA{} on the same split, the samples are effectively the same structure.
The cause is architectural: the top-level anchor flow is a deterministic ODE
started from the apo structure, so all 20 samples share identical anchors
(0.0~\AA{} difference), and stochasticity enters only through the midpoint
flow, which fills gaps between those fixed parents.}

\begin{table}[h]
\caption{\rev{Single sample versus the best of 20 under an oracle that selects
the sample closest to the true holo structure. Fixed 150-pair D3PM test split,
five seeds.}}
\label{tab:diversity}
\centering
\small
\revfit{%
\begin{tabular}{lcc}
\toprule
& Single sample & Best of 20 (oracle)\\
\midrule
Median RMSD (\AA) $\downarrow$ & 1.333 $\pm$ 0.056 & 1.330 $\pm$ 0.060\\
Mean RMSD (\AA) $\downarrow$ & 2.766 $\pm$ 0.041 & 2.756 $\pm$ 0.042\\
$<2$~\AA{} (\%) $\uparrow$ & 58.53 $\pm$ 0.73 & 58.80 $\pm$ 0.56\\
Dist.\ fidelity (\%) $\uparrow$ & 99.58 $\pm$ 0.02 & 99.58 $\pm$ 0.02\\
Angle fidelity (\%) $\uparrow$ & 95.87 $\pm$ 0.25 & 95.92 $\pm$ 0.25\\
MolProbity $\downarrow$ & 3.031 $\pm$ 0.052 & 3.024 $\pm$ 0.054\\
\bottomrule
\end{tabular}}
\end{table}

\rev{Even the oracle gives almost the same accuracy as a single sample.
\method{} in its current form therefore does not generate structural diversity
or alternative holo states: it is not an ensemble generator but a nearly
deterministic predictor mapping one apo structure to one holo-like structure.
This limitation is stated in Section~\ref{sec:discussion}.}

\end{document}